\documentclass[a4paper,fleqn]{cas-dc}

\usepackage[authoryear]{natbib}
\usepackage{enumitem}
\setlist{nosep, leftmargin=14pt}

\usepackage{mwe} 
\usepackage{algorithm}
\usepackage{algpseudocode}
\usepackage{amsmath} 
\usepackage{comment}
\usepackage{graphicx}
\usepackage{xcolor}
\usepackage[colorinlistoftodos]{todonotes}
\usepackage{hyperref}
\def\tsc#1{\csdef{#1}{\textsc{\lowercase{#1}}\xspace}}
\tsc{WGM}
\tsc{QE}

\begin{document}

\let\WriteBookmarks\relax
\def\floatpagepagefraction{1}
\def\textpagefraction{.001}

\shorttitle{ }
\shortauthors{ }

\title [mode = title]{CurvSegFlow: Time-Conditioned Flow Matching for Robust Segmentation of Curvilinear Structures in Noisy Biomedical Images}  



%

\author[1]{Sidi Mohamed Sid'El Moctar}
\cormark[1]
\ead{sidimohamed.sidelmoctar@univ-rennes.fr}
\credit{Conceptualization, Methodology, Software, Investigation, Formal analysis, Visualization, Writing – original draft, Writing – review and editing}

\author[1, 2]{Achraf Ait Laydi}
\credit{Formal analysis, Validation, Writing – review and editing}

\author[3]{Alexandre Beber}
\credit{Data curation, Validation, Writing – review and editing}

\author[3]{Marcus Braun}
\credit{Data curation, Validation}

\author[3]{Zdenek Lansky}
\credit{Data curation, Validation}

\author[2]{Yousef El Mourabit}
\credit{Validation, Writing – review and editing}

\author[1]{Hélène Bouvrais}
\cormark[1]
\ead{helene.bouvrais@univ-rennes.fr}
\credit{Conceptualization, Formal analysis, Data curation, Funding acquisition, Project administration, Resources, Supervision, Validation, Writing – review and editing}




\affiliation[1]{organization={CNRS, Univ. Rennes, Institute of Genetics and Development of Rennes (IGDR)},
            city={Rennes},
            country={France}}
            
\affiliation[2]{organization={TIAD Laboratory, Sciences and Technology Faculty, Sultan Moulay Slimane Univ},
            city={Beni Mellal},
            country={Morocco}}

\affiliation[3]{organization={Institute of Biotechnology, Czech Academy of Sciences},
            city={Vestec},
            country={Czech Republic}}








\begin{abstract}
Accurate segmentation of curvilinear structures remains challenging in biomedical imaging due to their thin geometry, complex topology, and sensitivity to noise. This is particularly critical for microscopy images of cytoskeletal network, where low signal-to-noise ratios and dense filament crossings often lead to fragmented or inaccurate segmentation. In this work, we propose \textit{CurvSegFlow}, a segmentation framework based on time-conditioned flow matching. Instead of predicting a segmentation mask in a single pass, the method models segmentation as a dynamic process that progressively refines a noisy initialization into the target structure through a learned velocity field.
The proposed model combines a U-Net backbone with triple-term loss function and temporal embeddings to guide the refinement process across reconstruction stages. This formulation enables gradual error correction and improves the continuity of thin structures. CurvSegFlow is evaluated on multiple synthetic and real microtubule datasets, as well as on public benchmarks of retinal vessels, corneal nerves and coronary arteries. Across datasets, the method achieves competitive or superior performance compared to established segmentation models, with consistent improvements in precision and structural continuity, particularly under low signal-to-noise conditions.
These results show that flow-based iterative refinement provides a robust and general framework for curvilinear structure segmentation. Overall, the proposed approach improves segmentation quality in challenging imaging conditions and generalizes effectively across modalities without architectural changes.
\end{abstract}
%


\maketitle

\section{Introduction}
\label{sec:intro}

Curvilinear structures are elongated, tubular, and often branching patterns that appear in a wide range of biomedical images. They include anatomical and cellular structures such as blood vessels, nerve fibers, and cytoskeletal filaments, whose geometry and topology are closely linked to biological function and disease progression \citep{kv2023segmentation}. Structural alterations in these networks are associated with numerous pathological conditions, including cardiovascular diseases, neurodegeneration, and cancer \citep{brunden2017altered}. As a result, accurate extraction of curvilinear structures is a fundamental step in quantitative image analysis and computer-aided diagnosis.
Among curvilinear structures, microtubules provide a representative and challenging example. As a major component of the cytoskeleton, microtubules play key roles, e.g. in intracellular transport, cell division, and morphogenesis \citep{bershadsky2012cytoskeleton}. They are highly dynamic filaments that continuously switch between growth and shrinkage, regulating cell shape and organization. Quantifying their spatial organization from microscopy images is essential for studying cellular processes and understanding diseases \citep{lafanechere2022microtubule, brunden2017altered}.

Curvilinear structures are observed across multiple imaging modalities, including fluorescence microscopy, fundus imaging, optical coherence tomography, and magnetic resonance angiography. Despite differences in acquisition techniques, these structures share common visual characteristics: they are thin, elongated, and often form complex, highly connected networks. This variability across modalities and scales makes the design of robust and generalizable segmentation methods particularly challenging. As a result, many existing tools developed are tailored to specific types of curvilinear structures. For example,  recent tools such as FAST  (for filamentous actin segmentation) \citep{aljapur2026fast} and PA-Net  (for retinal vessel segmentation) \citep{luo2025pa} were developed for specialized applications.

Accurate segmentation of curvilinear structures remains challenging due to multiple factors. Their apparent width often approaches the imaging resolution limit, making pixel-wise classification ambiguous. Additionally, these structures exhibit strong local intensity variations, frequent crossings, and complex spatial arrangements. In microscopy, additional challenges arise from photon noise, background fluorescence, uneven illumination, and low signal-to-noise ratios, the latter often worsened by short exposure times. Similar issues may also be encountered in other imaging modalities.
Last, preserving topology and connectivity is also a major challenge. While many deep learning models produce accurate pixel-wise predictions, they often fail to maintain connectivity, leading to fragmented structures. This limitation is particularly significant, for vessels, nerves, and filamentous structures where topology carries important biological information \citep{hu2019topology}.

Traditional methods for curvilinear structure extraction rely on hand-crafted features, such as ridge detectors, steerable filters, or graph-based tracing algorithms. While effective in controlled scenarios, these approaches require careful parameter tuning and do not generalize well across datasets \citep{kv2023segmentation}. Deep learning methods, particularly convolutional neural networks such as U-Net \citep{ronneberger2015u}, have become the standard approach for biomedical image segmentation by learning hierarchical feature representations. More recent architectures incorporate attention mechanisms or transformers to model long-range dependencies, though this comes at the cost of increased model complexity \citep{azad2024advances}. These methods also typically require large annotated datasets and may struggle to generalize across imaging domains. Critically, most existing methods predict segmentation masks in a single forward pass, which makes them sensitive to early mistakes, especially for thin and low-contrast structures, resulting in broken or discontinuous predictions.

In contrast, modeling segmentation as a continuous transformation allows predictions to evolve gradually. Each step refines the previous estimate, which helps correct local errors rather than propagating them.
This idea is inspired by recent generative approaches such as diffusion models, which are iterative refinement methods \citep{kazerouni2023diffusion}. However, diffusion models rely on stochastic sampling and repeated noise removal, which can introduce blur and degrade fine structures when many steps are required \citep{kazerouni2023diffusion}.
Flow matching provides an alternative formulation. It learns a deterministic vector field that directly transports an initial distribution to the target through a continuous trajectory \citep{lipman2022flow}. This avoids stochastic sampling, enabling more stable and efficient inference \citep{bogensperger2025flowsdf}.
Importantly, this continuous evolution acts as implicit regularization: the segmentation evolves smoothly over time, which naturally limits abrupt structural inconsistencies. This is particularly relevant for curvilinear structures.

While generative approaches, including diffusion and flow-based models, have recently been explored for medical image segmentation, their application to curvilinear structures remains limited \citep{kazerouni2023diffusion}. Diffusion-based methods have shown promising results, particularly in low-data regimes \citep{wu2024medsegdiff}, but they typically require many iterative steps and high computational cost. Recent works, such as FlowSDF \citep{bogensperger2025flowsdf}, PolypFlow \citep{wang2025polypflow} and rectified-flow formulations \citep{wang2024semflow, schusterbauer2025diff2flow}, highlight the potential of flow-based modeling for segmentation tasks.
More recently, FMS$^2$ \citep{asadi2026fms} introduced a unified flow matching framework for segmentation and synthesis of thin structures, showing that flow-based formulations can better capture fine-scale geometry. On the  DRIVE dataset, FMS$^2$ achieved an Dice score of 0.789. However, existing approaches mainly focus on joint generation and segmentation or global shape modeling. They do not explicitly address the progressive refinement of noisy segmentations or its impact on preserving connectivity in low signal-to-noise conditions. This focus is critical for curvilinear structures, where small local errors can lead to significant topological inconsistencies \citep{mou2021cs2}.

In this work, we introduce \textit{CurvSegFlow}, a flow-matching-based framework for curvilinear structure segmentation. We indeed investigate conditional flow matching as an image-guided mechanism for progressively refining curvilinear segmentation \citep{tong2023conditional}. Instead of treating it as a generative modeling approach, we use it to iteratively correct a mask conditioned on the input image, starting from a noisy initialization and learning a time-dependent velocity field that drives the prediction toward the target. Unlike traditional methods that model the global distribution of segmentation masks, CurvSegFlow formulates segmentation as an explicit iterative refinement process, in which a learned flow progressively corrects errors and restores connectivity from a noisy initial mask. This formulation enables the model to learn gradual refinement dynamics rather than relying on a single-step prediction. It is particularly well-suited for curvilinear structures, as it improves robustness to noise while preserving structural continuity across diverse imaging conditions. To further improve both trajectory quality and final segmentation accuracy, we combine the flow-matching objective with standard segmentation losses. We therefore implement a dual-loss control: (1) a Mean Square Error (MSE) loss guides the trajectory reconstruction (vector field estimation), and (2) Dice and Weighted Binary Cross Entropy (BCE) losses align the final predicted mask with the ground truth.

CurvSegFlow extends our prior work, MTFlow \citep{sid2026mtflow}, which introduced flow matching for microtubule segmentation using a single weighted BCE loss. Key improvements include (i) implementing a triple-term loss function, (ii) incorporating attention gates in skip connections to better focus on thin and low-contrast structures, (iii) demonstrating stronger cross-domain generalization across diverse curvilinear benchmarks beyond microtubules, and (iv) comparing CurvSegFlow with a wide range of recent deep-learning models, showcasing its state-of-the-art perfromance. Additionally, two ablation studies highlight how CurvSegFlow’s implementation surpasses the initial implementation of MTFlow, and the optimal number of integration steps for inference is investigated. We validate CurvSegFlow on both synthetic and real datasets of microtubules and further assess its generalization on public benchmarks of curvilinear structures, including microtubules, retinal vessels (DRIVE and CHASEDB1), corneal nerves (CORN-1), and coronary arteries (ARCADE). 

The main contributions of this work are:
\begin{itemize}
    \item We investigate conditional flow matching as a continuous refinement framework for curvilinear structure segmentation, with particular emphasis on challenging low signal-to-noise conditions.
    \item We propose a time-conditioned architecture that iteratively refines segmentation masks via learned vector fields, beginning with a noisy initialization and guided by a dual-loss control.
    \item We provide a comprehensive evaluation across microscopy, vascular, and nerve imaging datasets, demonstrating consistent performance and strong cross-domain generalization without task-specific architectural modifications.
\end{itemize}

\section{Methodology}
\label{sec:meth}

\subsection{Datasets}

\begin{figure*}
\centering
\includegraphics[width=0.99\linewidth]{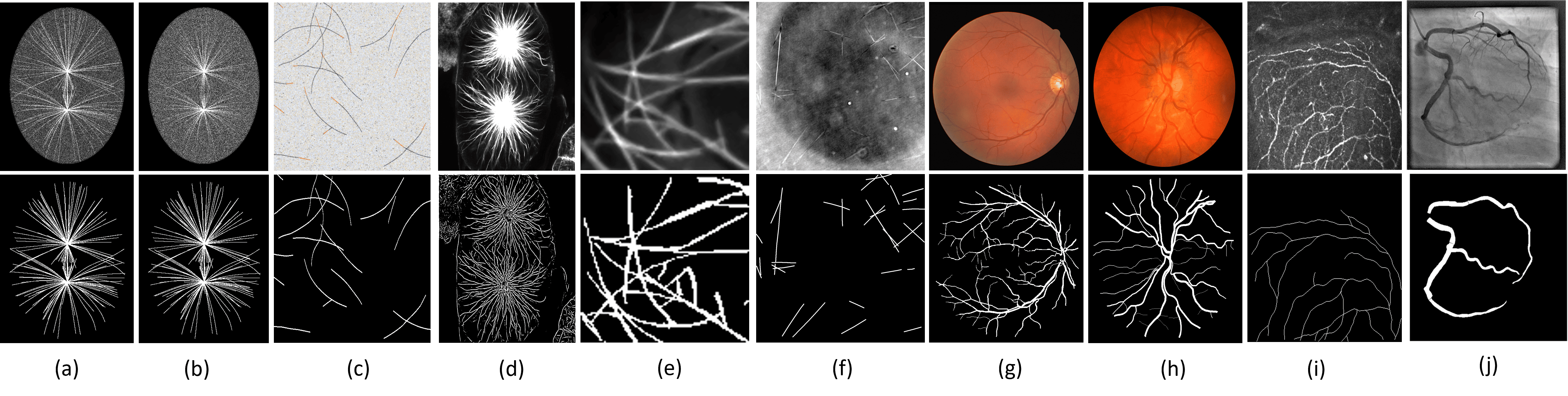}
\caption{\label{fig:datasets} \footnotesize Diversity of curvilinear structures and imaging modalities across the studied datasets.(Top row)  Exemplar images, and (bottom row) their corresponding annotations. From the left to right: MicSim\_FluoMT-Simple, MicSim\_FluoMT-Complex, SynthMT, MicReal\_FluoMT, Revised Higaki 2024, IRM\_InVitroMT, DRIVE, CHASEDB1, CORN-1, and ARCADE}
\end{figure*}

\subsubsection{Synthetic microtubule datasets}

To address the challenges of segmenting microtubules in fluorescence microscopy images, we used these public datasets:

\textbf{MicSim\_FluoMT} \citep{bouvrais2025micsim_fluomt}. These datasets were generated through a two-step pipeline designed to produce realistic confocal microscopy images of microtubule networks in \textit{Caenorhabditis elegans} zygotes, including fluorescence imaging artefacts, with perfectly aligned ground-truth masks.
First, Cytosim \citep{nedelec2007collective} simulated astral microtubule networks resembling those of \textit{C. elegans} zygotes in mitosis. Biologically informed parameters were employed to generate diverse filament geometries. Second, ConfocalGN converted these simulations into confocal-like fluorescence images by modelling optical blur, photon noise, and empirically derived background fluorescence intensities \citep{dmitrieff2017confocalgn}. Binary masks were directly obtained from the simulations to ensure precise annotation, even at low signal-to-noise ratios.
The pipeline produced two datasets with different segmentation difficulty levels. In \textbf{MicSim\_FluoMT-Simple} (Figure \ref{fig:datasets}-a), filaments have uniform fluorescence intensity, whereas in \textbf{MicSim\_FluoMT-Complex} (Figure \ref{fig:datasets}-b), intensity decreases toward filament extremities, making segmentation more challenging. Each dataset contains 1192 images (666 $\times$ 666 pixels) with variability in filament density and morphology. For both variants, 953 images were used for training, 119 for validation, and 120 for testing.

\textbf{SynthMT} \citep{koddenbrock2026synthetic}: This recent dataset contains 6600 synthetic interference reflection microscopy (IRM) images of microtubules (512 $\times$ 512 pixels) with low background noise (Figure \ref{fig:datasets}-c). To maintain consistency with our training protocol and computational limits, we randomly selected a subset of 400 images from the full dataset. This subset was split into 100 images for training, 150 for validation, and 150 for testing.

\subsubsection{Real microtubule datasets}

To evaluate performance on real microscopy data from various modalities (different microscopies, distinct cell types or microtubule labelings), we used these three datasets :

\textbf{MicReal\_FluoMT dataset} \citep{cueff2025micreal_fluomt}: It contains 49 images (1032 $\times$ 1032 pixels) of stained microtubules acquired from \textit{C. elegans} embryos using Airyscan 2 confocal microscope (Figure \ref{fig:datasets}-d). Images were captured under multiple experimental conditions to promote variability in microtubule morphology and density. Segmentation masks were generated using a semi-automated three-step pipeline. First, extended depth-of-field projections were computed to improve filament continuity. Second, filament-enhancing orientation filtering was applied to increase contrast relative to background noise. Finally, interactive machine-learning segmentation was performed using Ilastik, where representative regions were manually annotated to train a classifier that was applied across the dataset. The resulting masks may include occasional non-microtubule structures due to staining artefacts or segmentation limitations.
The dataset was split into 29 images for training, 10 for validation, and 10 for testing. 

\textbf{Higaki 2024} \citep{guo2025synseg}: An independent microtubule segmentation dataset containing confocal images of \textit{Nicotiana tabacum} BY-2 cells (100 $\times$ 100 pixels) labelled with YFP-tubulin was originally released in 2024 \citep{horiuchi2024deep}. Its ground-truth masks were generated using semi-automated thresholding and morphological filtering. The test split of this dataset was later manually reannotated to provide a more accurate and consistent reference standard \citep{guo2025synseg}. The revised Higaki 2024 dataset now includes 98 annotated images (Figure \ref{fig:datasets}-e), used for 5-fold cross-validation in segmentation evaluation.

\textbf{IRM\_InVitroMT} \citep{Beber2026IRM}: A new microtubule dataset of interference microscopy images was provided in this work, where microtubules were purified from porcine brain, polymerized, and stabilized with Taxol and/or GMPCPP following established protocols \citep{siahaan2022microtubule}. For interference microscopy, glass coverslips were plasma cleaned, silanized, functionalized with anti-tubulin antibodies, and passivated with pluronic F127 before introducing microtubules. Images were acquired using four microscopes (two Nikon Eclipse Ti-E and two Nikon Eclipse Ti-2) with  60$\times$ or 100$\times$ oil objectives (NA 1.49) Illumination was provided by either a CoolLED pE-300 or Nikon Intenslight E light source. Images were collected over several years and manually selected to ensure dataset diversity based on illumination patterns, signal-to-noise ratios, and microtubule coverage, size and orientation. Microtubules were manually annotated in ImageJ using the freehand line tool (line thickness: 3 pixels for 60x objective, or 5 pixels for 100x objective). Annotations were converted to binary masks via a custom ImageJ macro, and ground-truth images were inverted to account for the bright field imaging modality. The IRM\_InVitroMT dataset comprises 237 images (original sizes: 706 $\times$ 644 to 2048 $\times$ 2048 pixels), resized to 1280 $\times$ 1280 pixels (Figure \ref{fig:datasets}-f). The dataset was split into 206 training images and 31 test images.

\subsubsection{Additional curvilinear structure datasets}
To test the generalization performance to other curvilinear structures, experiments were conducted on four publicly available datasets: DRIVE, CHASEDB1, CORN-1, and ARCADE. These datasets are widely used benchmark for segmenting blood vessels, nerves and coronary arteries, offering diverse imaging characteristics and anatomical structures for comprehensive validation.

\textbf{DRIVE} \citep{staal2004ridge}: This dataset consists of 40 color fundus images (565 $\times$ 584 pixels) acquired using a Canon CR5 non-mydriatic 3-CCD camera with a $45^\circ$ field of view (Figure \ref{fig:datasets}-g). The dataset was originally divided into 20 training images and 20 testing images, and this predefined partition is followed in our experiments. Expert manual annotations of blood vessels serve as ground truth for segmentation evaluation.

\textbf{CHASEDB1} \citep{fraz2012ensemble}: It contains 28 retinal fundus images (999 $\times$ 960 pixels) captured from both eyes of 14 school children using a hand-held Nidek NM-200-D fundus camera (Figure \ref{fig:datasets}-h). Two independent experts manually annotated the retinal blood vessel structures for each image. Since predefined masks are not included, segmentation masks were derived from the provided expert annotations and used as ground truth in this study. The dataset was split into 20 images for training, 3 for validation and 5 for test.

\textbf{CORN-1} \citep{imedningbo}: It is a publicly available corneal confocal microscopy dataset comprising 1698 images (384 $\times$ 384 pixels) of corneal subbasal nerve fibers (Figure \ref{fig:datasets}-i). The images were acquired using a Heidelberg Retina Tomograph III equipped with a Rostock Cornea Module microscope, and collected by the Peking University Third Hospital, China, and the University of Padova, Italy, ensuring clinical diversity. Manual annotations traced by ophthalmologists using ImageJ provide accurate ground truth for nerve fiber segmentation. A 5-fold cross-validation was used for segmentation evaluation.

\textbf{ARCADE} \citep{popov2024dataset}: The Automatic Region-based Coronary Artery Disease Diagnostics using X-ray angiography imagEs (ARCADE) dataset is a publicly available benchmark for coronary vessel analysis from X-ray coronary angiography images (512 $\times$ 512 pixels) (Figure \ref{fig:datasets}-j). It provides pixel-level annotations of the coronary arterial tree following the SYNTAX score protocol, which divides the vasculature into 26 anatomically defined segments, enabling detailed modeling of vessel topology, bifurcations, and caliber variations. In this work, we focus on vessel segmentation and adopt the standard split of 1000 training and 200 validation images, along with an independent test set of 300 images for evaluation.

\subsection{CurvSegFlow model}

\begin{figure*}
\vspace{-10pt}
\centering
\includegraphics[width=0.99\linewidth]{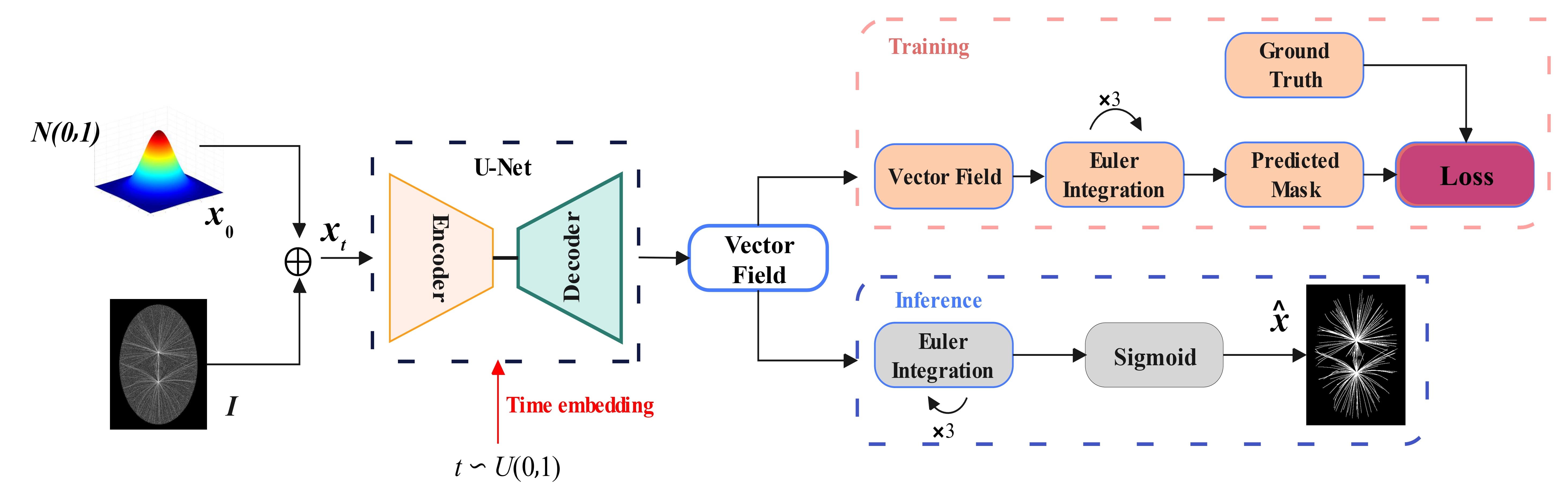}
\caption{\label{fig:mt_flow} \footnotesize An overview of the proposed CurvSegFlow model}
\end{figure*}

We propose CurvSegFlow, a segmentation framework that formulates curvilinear stucture segmentation as a dynamic reconstruction process. Instead of directly predicting a binary mask, the model learns a time-dependent velocity field that progressively transforms a noisy initialization into the ground-truth segmentation as shown in Figure \ref{fig:mt_flow}.
Given an input image $I$, which we aim to segment, the model operates on an input formed by concatenating the image $I$ with a noisy initialization. RGB images contribute three channels, while grayscale images are treated as a single channel. In both cases, an additional channel corresponds to the evolving mask, resulting in a consistent input structure.

We begin by sampling a noisy initialization $x_0 \in \mathbb{R}^{H \times W}$ from a standard normal distribution:
\begin{equation}
x_0 \sim \mathcal{N}(0,1),
\end{equation}
where $H$ and $W$ denote the spatial dimensions of the input image $I$. 

Let $x_1 \in \{0,1\}^{H \times W}$ be the ground-truth binary mask.
A continuous interpolation between $x_0$ and $x_1$ is defined as:
\begin{equation}
x_t = (1 - t)x_0 + t x_1, \quad t \in [0,1],
\end{equation}
where $t$ is a scalar time variable sampled uniformly from the interval $[0,1]$. The variable $x_t$ represents an intermediate state between noise and the true segmentation, also named as the evolving mask. 

The objective is that the model learns a velocity field $v_\theta(x_t, I, t)$, parameterized by $\theta$, that predicts how $x_t$ should evolve with respect to time. The target velocity is defined as:
\begin{equation}
u_t = x_1 - x_0,
\end{equation}
which corresponds to the optimal transport direction between the initial noise and the ground-truth mask. This velocity is constant along the interpolation path.

\begin{figure*}
\centering
\includegraphics[width=0.8\linewidth, height = 7cm]{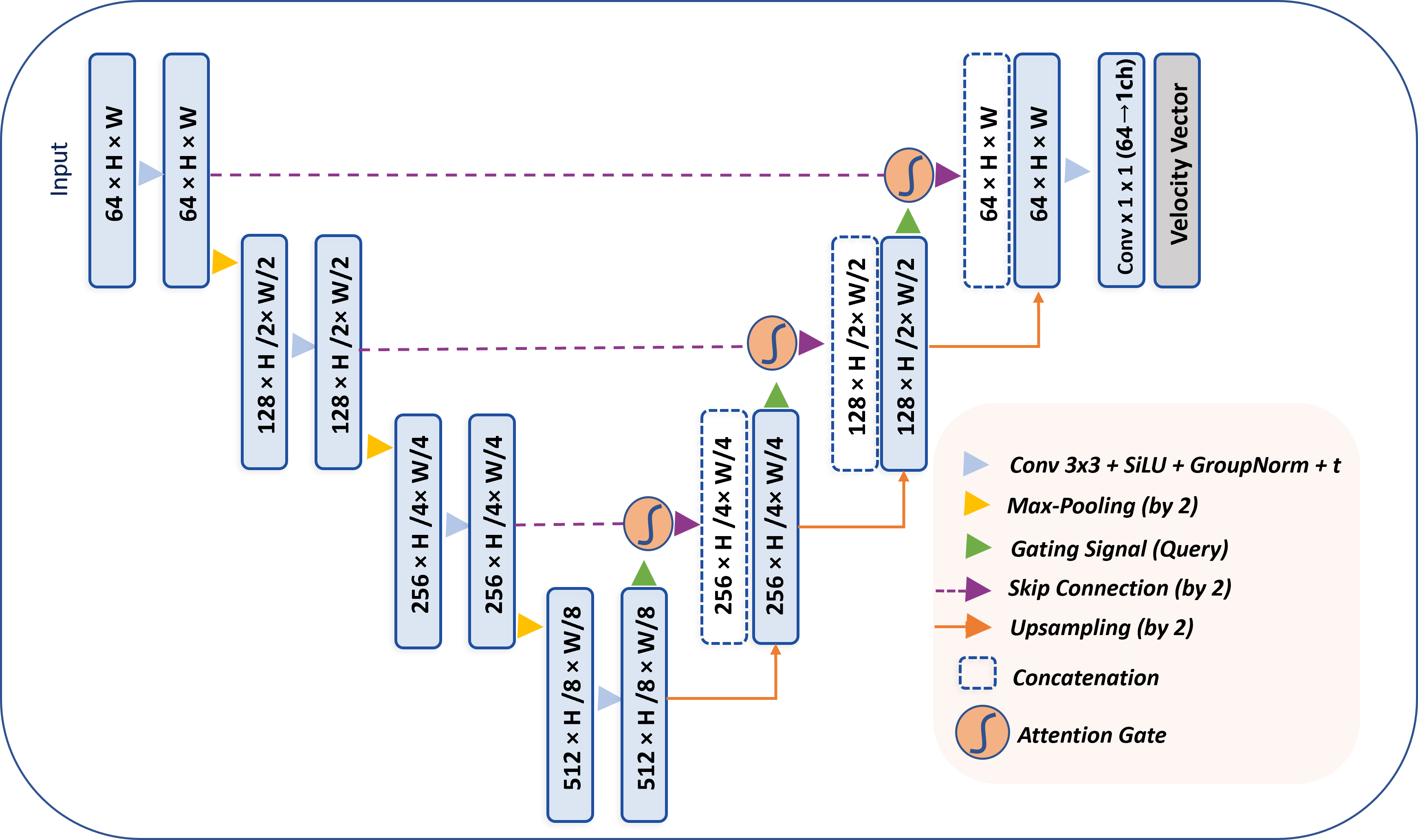}
\caption{\label{fig:unet_flow} \footnotesize Time-conditioned U-Net architecture of CurvSegFlow}
\end{figure*}

The network architecture is based on a time-conditioned U-Net \citep{ho2020denoising} as shown in Figure \ref{fig:unet_flow}. The encoder consists of four resolution levels with feature dimensions $64$, $128$, $256$, and $512$. Each level is composed of two convolutional layers with kernel size $3 \times 3$, followed by Group Normalization with 8 groups and SiLU activation functions. Down-sampling is performed using max-pooling operations.

Time information is encoded using sinusoidal embeddings \citep{vaswani2017attention, ho2020denoising}. The scalar time $t$ is mapped to a higher-dimensional representation using sine and cosine functions. This embedding is then processed by a small multi-layer perceptron and added to the feature maps inside each convolutional block. This conditioning allows the network to adapt its behavior depending on the reconstruction stage.
The decoder mirrors the encoder structure. It uses transposed convolutions for up-sampling and progressively reconstructs spatial details. At each resolution level, encoder features are combined with decoder features through skip connections.

\begin{figure}
\centering
\includegraphics[width=0.9\linewidth]{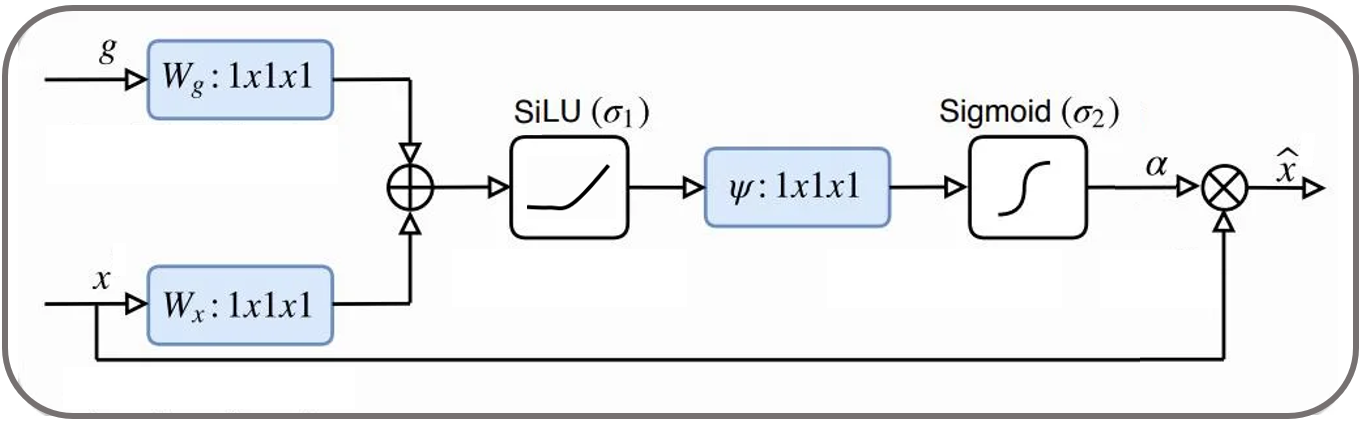}
\caption{\label{fig:att_gate} \footnotesize The attention gate}
\end{figure}

To improve feature selection, attention gates \citep{oktay2018attention}, detailed in Figure \ref{fig:att_gate} are applied to all skip connections. Each gate takes two inputs: a gating signal $g$ from the decoder and a feature map $x$ from the encoder. Both signals are projected to a lower-dimensional space using $1 \times 1$ convolutions and normalized. They are then combined and passed through a non-linear activation followed by a sigmoid function to produce an attention map $\alpha \in (0,1)^{H \times W}$. The encoder features are scaled as:
\begin{equation}
\tilde{x} = \alpha \cdot x,
\end{equation}
where $\cdot$ denotes element-wise multiplication. This mechanism reduces irrelevant background regions and highlights structures of interest.

The final layer of the network is a $1 \times 1$ convolution that outputs a single-channel velocity field:
\begin{equation}
v_\theta(x_t, I, t) \in \mathbb{R}^{H \times W}.
\end{equation}
No activation function is applied at this stage, allowing the model to predict unbounded velocity values.
At inference time, the segmentation is obtained by integrating the learned dynamics over time using an explicit Euler scheme. Starting from $x_0$, the state is updated as:
\begin{equation}
x_{n+1} = x_n + \Delta t \, v_\theta(x_n, I, t_n),
\end{equation}
where $\Delta t = \frac{1}{N}$ is the step size, $N$ is the number of integration steps, and $t_n = n \Delta t$.

After $N$ steps, the final prediction is obtained by applying a sigmoid function:
\begin{equation}
\hat{x} = \sigma(x_N),
\end{equation}
where $\hat{x} \in (0,1)^{H \times W}$ represents the probability map of the segmentation.
This iterative process is expected to enable progressive refinement of the mask, allowing the model to correct errors, and recover fine structures, and improve robustness to noise and variations in appearance.

From a modeling perspective, the proposed formulation can be interpreted as learning a transport map between a simple prior (noise) and the target segmentation mask. Unlike classical discriminative models that directly predict the final output, the model learns how to update the current estimate at each step. This perspective is closely related to continuous normalizing flows, where a transformation is defined through an ordinary differential equation parameterized by a neural network \citep{chen2018neural, grathwohl2018ffjord}. In our case, this formulation is adapted to segmentation and conditioned on the input image, which allows the model to integrate both spatial context and temporal refinement \citep{po2024state}.

\subsection{Training Protocol}

The model is trained using a conditional flow matching strategy, summarized in Algorithm \ref{alg:CurvSegFlow}. For each training sample, a time value $t \sim \mathcal{U}(0,1)$ and an initial noise $x_0 \sim \mathcal{N}(0,1)$ are sampled. The interpolated state $x_t$ is computed and used as input to the network together with the image $I$.
The model predicts a velocity field $v_\theta(x_t, I, t)$, which is compared to the target velocity $u_t = x_1 - x_0$ using a mean squared error loss:
\begin{equation}
\mathcal{L}_{\text{FM}} = \frac{1}{2} \| v_\theta(x_t, I, t) - u_t \|^2.
\end{equation}

A short differentiable Euler integration with a small number of steps is performed to obtain a predicted probability map $\hat{x}$. This allows gradients to propagate through the reconstruction process.

In addition to the velocity loss, a segmentation loss is applied on the reconstructed mask. To handle class imbalance between foreground and background pixels, a weighted binary cross-entropy loss is used:

\begin{equation}
\mathcal{L}_{\text{WBCE}} =
- \frac{1}{HW} \sum_{i,j} \big[ w_1 y_{i,j} \log \hat{y}_{i,j} + w_0 (1-y_{i,j})\log(1-\hat{y}_{i,j}) \big]
\end{equation}

where $y_{i,j}$ and $\hat{y}_{i,j}$ denote the ground-truth and predicted values at pixel $(i,j)$, and $w_1$ and $w_0$ are the weights for foreground and background classes.

A soft Dice loss is also included to directly optimize overlap between prediction and ground truth:
\begin{equation}
\mathcal{L}_{\text{Dice}} =
1 - \frac{2 \sum_{i,j} y_{i,j} \hat{y}_{i,j} + \epsilon}
{\sum_{i,j} y_{i,j} + \sum_{i,j} \hat{y}_{i,j} + \epsilon},
\end{equation}
where $\epsilon$ is a small constant to ensure numerical stability.

The total loss is defined as a weighted combination:
\begin{equation}
\mathcal{L} =
\lambda_{\text{FM}} \mathcal{L}_{\text{FM}} +
\lambda_{\text{WBCE}} \mathcal{L}_{\text{WBCE}} +
\lambda_{\text{Dice}} \mathcal{L}_{\text{Dice}},
\end{equation}
where $\lambda_{\text{FM}}$, $\lambda_{\text{WBCE}}$, and $\lambda_{\text{Dice}}$ control the contributions of each term.

The training is performed using the AdamW optimizer with a learning rate of $1 \times 10^{-4}$ and a weight decay of $1 \times 10^{-5}$. A cosine annealing schedule is used to gradually decrease the learning rate over time. 
Early stopping is employed based on the validation loss, with a fixed patience to prevent overfitting. The model achieving the best validation performance is saved for final evaluation. 
Data augmentation includes random horizontal and vertical flips, as well as random rotations within $\pm 15^\circ$, applied jointly to images and masks. 

Due to its iterative nature, CurvSegFlow requires multiple forward passes during training. As a result, it is slower than some lightweight one-shot models. However, the number of refinement steps can be adjusted to balance accuracy and speed. This trade-off is similar to diffusion-based models, which rely on iterative refinement and therefore incur higher computational cost. However, unlike standard diffusion processes that require long stochastic denoising chains, CurvSegFlow learns a deterministic transport field with approximately linear dynamics between noise and target masks. This makes the underlying trajectory significantly easier to integrate, allowing accurate reconstruction with relatively few refinement steps. Similar observations have been made in flow-based and rectified flow models, where learning straightened probability paths leads to fast convergence and reduced sampling steps \citep{lipman2022flow, song2023consistency}. In practice, we observe that CurvSegFlow achieves strong performance with a small number of integration steps, consistent with recent findings in continuous generative modeling that high-quality solutions can be obtained with few-step ODE solvers \citep{chen2018neural, grathwohl2018ffjord}. 

\begin{algorithm}[t]
\footnotesize
\caption{CurvSegFlow Training}
\label{alg:CurvSegFlow}
\begin{algorithmic}
\State \textbf{Input:} Training images $I$, ground-truth masks $x_1$, epochs $E$
\State \textbf{Initialize:} Model $v_\theta$, AdamW optimizer, learning rate scheduler
\State \textbf{Set:} $\lambda_{\text{FM}}, \lambda_{\text{WBCE}}, \lambda_{\text{Dice}}$

\For{epoch $=1$ to $E$}
    \For{each batch $(I, x_1)$}
        \State Sample $t \sim \mathcal{U}(0,1)$, $x_0 \sim \mathcal{N}(0,1)$
        \State $x_t = (1-t)x_0 + t x_1$, \quad $u_t = x_1 - x_0$
        
        \State $\hat{v} \leftarrow v_\theta(x_t, I, t)$
        \State $\mathcal{L}_{\text{FM}} = \frac{1}{2}\|\hat{v} - u_t\|^2$
        
        \State Initialize $x \leftarrow x_0$
        \For{$k = 1$ to $K$}
            \State $t_k = \frac{k}{K}$
            \State $\hat{v} \leftarrow v_\theta(x, I, t_k)$
            \State $x \leftarrow x + \Delta t \, \hat{v}$ \quad where $\Delta t = \frac{1}{K}$
        \EndFor
        
        \State $\hat{x} = \sigma(x)$
        
        \State Compute $\mathcal{L}_{\text{WBCE}}(\hat{x}, x_1)$
        \State Compute $\mathcal{L}_{\text{Dice}}(\hat{x}, x_1)$
        
        \State $\mathcal{L} = \lambda_{\text{FM}} \mathcal{L}_{\text{FM}} 
        + \lambda_{\text{WBCE}} \mathcal{L}_{\text{WBCE}} 
        + \lambda_{\text{Dice}} \mathcal{L}_{\text{Dice}}$
        
        \State Update $\theta$
    \EndFor
    
    \State Update scheduler; evaluate on validation set
\EndFor

\State \textbf{Output:} $v_\theta$
\end{algorithmic}
\end{algorithm}

\section{Results}
\label{sec:result}
To assess the performance of CurvSegFlow in curvilinear structure segmentation, we compared it with several widely used deep-learning models for biomedical image segmentation, as very few methods were specifically designed for this task. The selected models included U-Net \citep{ronneberger2015u}, U-Net++ \citep{zhou2018unet}, and ResUNet \citep{zhang2018road}, which have previously been used for segmenting curvilinear structures, as well as TransUNet \citep{chen2021transunet}, nnU-Net \citep{isensee2021nnu}  and CAR-UNet \citep{guo2021channel}, the latter specifically developed for blood vessel segmentation. All models shared the same backbone as CurvSegFlow, consisting of four down-sampling and up-sampling blocks. We also compared the model with MedSegDiff \citep{wu2024medsegdiff}, a diffusion probabilistic model, which has been developed for medical image segmentation, and SynSeg \citep{guo2025synseg}, which provides a pre-trained model dedicated to cytoskeleton segmentation.
CurvSegFlow, U-Net, U-Net++ and ResUNet were initialized with 64 filters, whereas TransUNet and CAR-UNet started with fewer filters (16 and 32 respectively) to reduce computational load. Segmentation performance was evaluated both qualitatively and quantitatively. Qualitative evaluation involved visual comparisons between prediction and ground-truth masks. Quantitative evaluation employed a range of metrics. On the one hand, Dice coefficient, sensitivity, precision, Matthews Correlation Coefficient (MCC), and area under the precision-recall curve (AUC-PR) were chosen, since they are reliable under severe class imbalance. On the other hand, Intersection over Union (IoU), specificity, accuracy, and Area Under the Curve (AUC-ROC) were  also computed, since they are classic metrics enabling comparisons with previous works. To specifically evaluate the preservation of microtubule topology and connectivity, two skeleton-based metrics were introduced: the centerline Dice (clDice) \citep{shit2021cldice}, which quantifies the overlap of predicted and ground truth centerlines, and the 95th percentile Hausdorff Distance (HD95), which provides a robust measure of boundary accuracy, which is insensitive to outliers.

\subsection{segmentation of microtubules}
\begin{figure*}[H]
    \centering
    \includegraphics[width=0.99\linewidth]{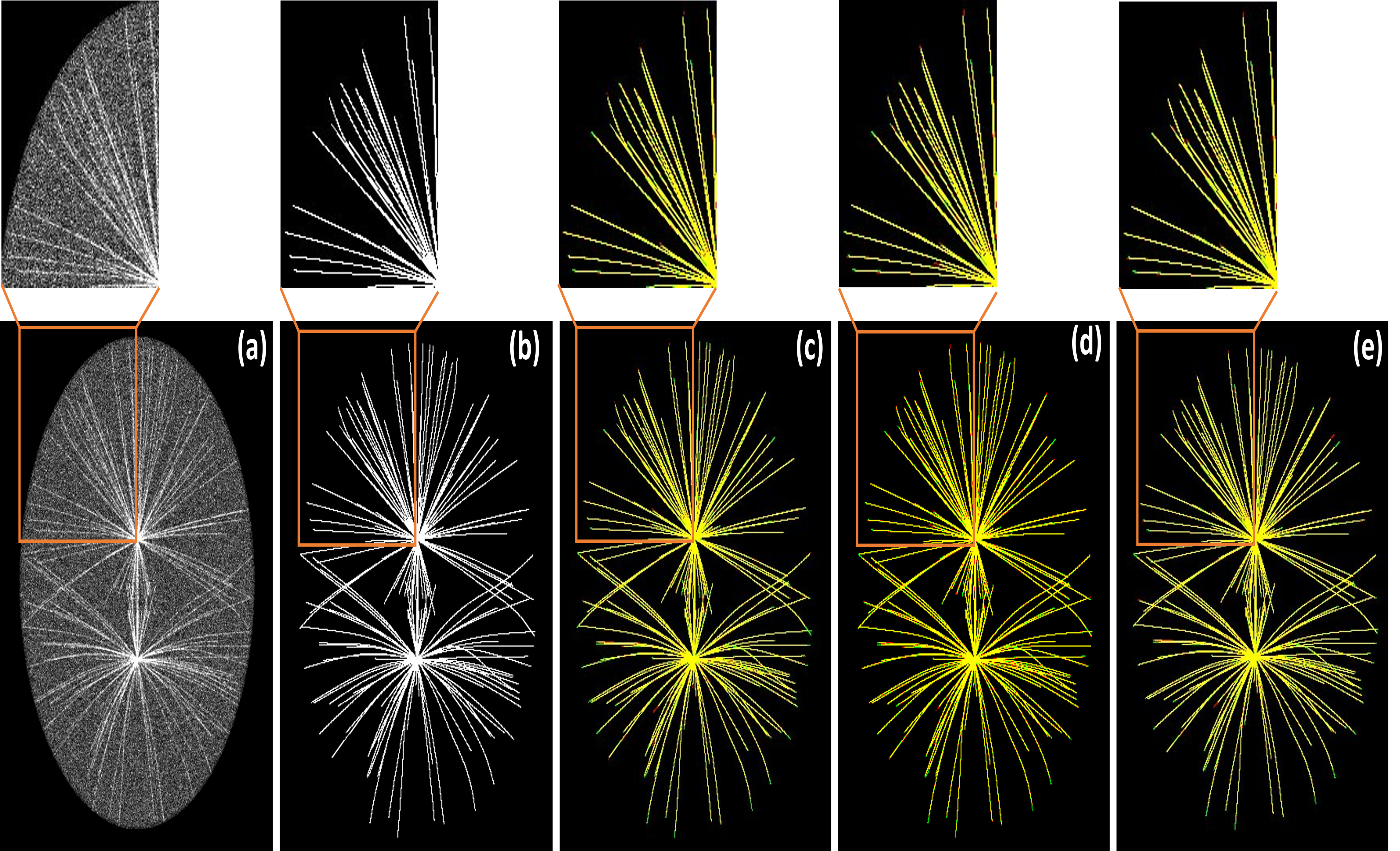}
    \caption{Segmentation on the MicSim\_FluoMT-Simple dataset: (a) Test image, (b) Ground truth, and (c-e) composite predictions from (c) CurvSegFlow, (d) U-Net++, and (e) nnU-Net, where true positives are shown in yellow, false positives in red, and false negatives in green.}
    \label{fig:seg_easy}
\end{figure*}

We first evaluated how CurvSegFlow performs in segmenting microtubules in synthetic images that include background fluorescence noise. 
On the MicSim\_FluoMT-Simple dataset, where fluorescence is visible along the entire length of the microtubules, all models achieved high Dice scores, indicating this task is not particulary challenging under these favorable signal conditions (Table~\ref{tab:simple}). Under these conditions, CurvSegFlow provided a consistent improvement.
CurvSegFlow achieved the highest overall performance, slightly outperforming nnU-Net and clearly surpassing other six models. This improvement was primarily driven by its strong precision, which was notably higher than that of the other models, and slightly higher than nnU-Net. Sensitivity remained high, though it was not the maximum compared to U-Net, U-Net++ and ResUnet. As shown in Figure ~\ref{fig:seg_easy}, false positives along background fluorescence are reduced compared to U-Net++, while filament continuity was preserved. These results demonstrated CurvSegFlow’s robustness in handling background fluorescence noise while maintaining high precision for segmenting tiny structures.

\begin{table*}
\centering
\footnotesize
\setlength{\tabcolsep}{3pt} 
\caption{Performance on the MicSim\_FluoMT-Simple dataset (mean across test images)}
\label{tab:simple}
\begin{tabular}{lccccccccccc}
\toprule
Model & Dice & IoU & SE & Prec. & SP & Acc. & MCC & AUC-ROC & AUC-PR & clDice & HD95 \\
\midrule
U-Net 2015 \citep{ronneberger2015u}
& 0.926 & 0.862 & \textbf{0.981} & 0.876 & 0.994 & 0.995 & 0.924 & \textbf{0.9995} & \textbf{0.9890} & \textbf{0.9532} & 1.000  \\

U-Net++ 2018 \citep{zhou2018unet}
& 0.924 & 0.859 & 0.981 & 0.874 & 0.994 & 0.995 & 0.923 & \textbf{0.9995} & 0.9880 & 0.9527 & 1.000  \\

ResUnet 2018 \citep{li2019residual}
& 0.9230 & 0.857 & 0.978 & 0.874 & 0.994 & 0.995 & 0.9213 & \textbf{0.9994} & 0.9871 & 0.9501 & 1.000  \\

CAR-Unet 2021 \citep{guo2021channel}
& 0.8852 & 0.7949 & 0.9587 & 0.8240 & 0.9917 & 0.9905 & 0.8836 & 0.9984 & 0.9688 & 0.9293 & 1.290  \\

TransUNet 2021 \citep{chen2021transunet}
& 0.844 & 0.730 & 0.969 & 0.748 & 0.987 & 0.990 & 0.845 & 0.998 & 0.942 & 0.9013 & 1.000  \\

nnU-Net 2021 \citep{isensee2021nnu}
& \textbf{0.9380} & \textbf{0.883} & 0.933 & 0.9435 & \textbf{0.9980} & \textbf{0.9950} & \textbf{0.9360} & 0.965 & 0.940 & 0.9521 & 1.000  \\

MedSegDiff 2024 \citep{wu2024medsegdiff}
& 0.8955 & 0.810 & 0.8945 & 0.8964 & 0.9958 & 0.9919 & 0.8913 & 0.9038 & 0.8166 & 0.9135 & 1.000  \\

\textbf{CurvSegFlow (Ours)} 
& 0.9350
& 0.8779
& 0.9247 
& \textbf{0.9453}
& \textbf{0.9980}
& \textbf{0.9950}
& 0.9324
& \textbf{0.9994}
& 0.9857
& 0.9484
& 1.000 \\
\bottomrule
\end{tabular}
\end{table*}

\begin{figure*}[H]
    \centering
    \includegraphics[width=0.99\linewidth]{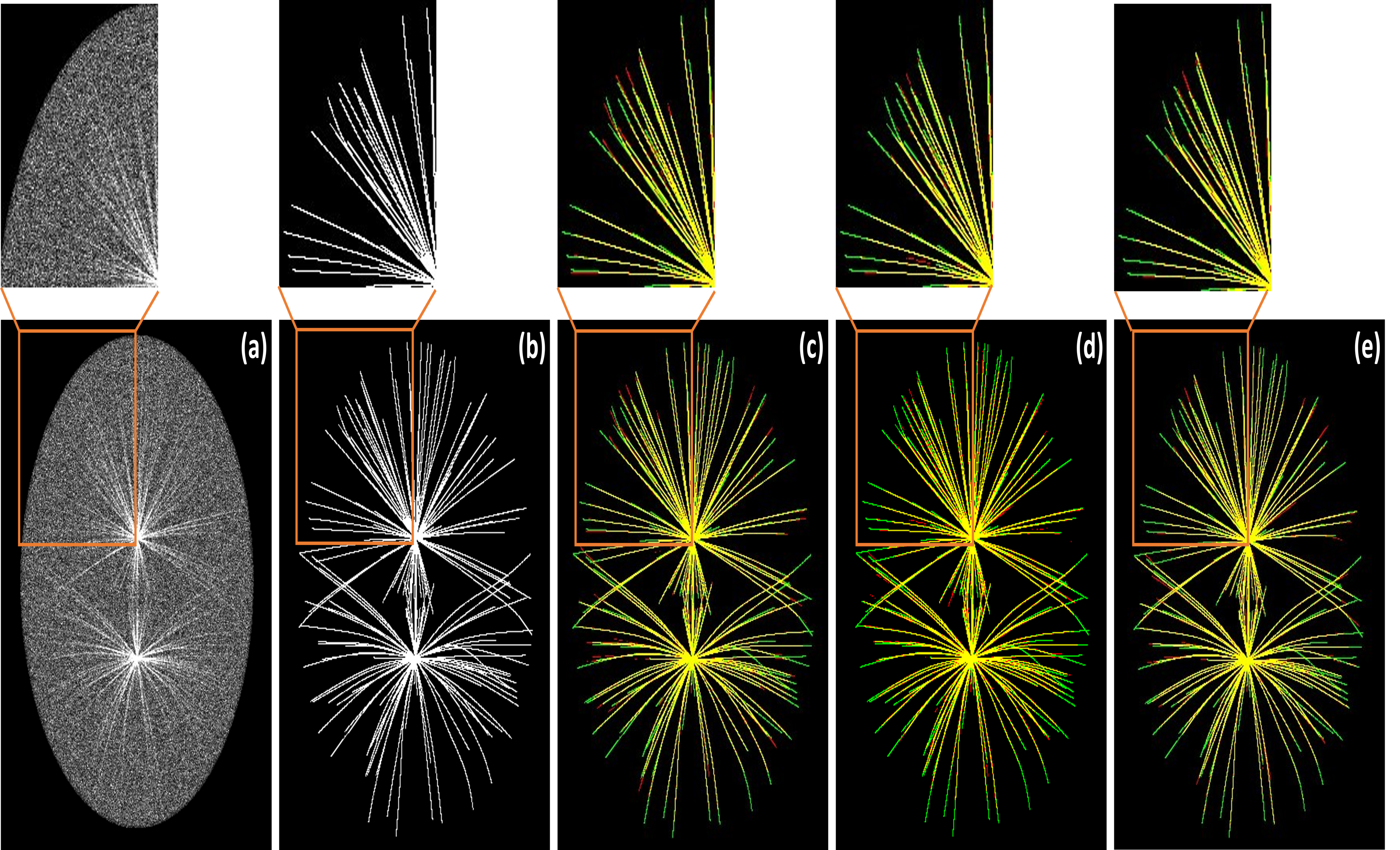}
    \caption{Segmentation on the MicSim\_FluoMT-Complex dataset: (a) Test image, (b) Ground truth, and (c-e) composite predictions from (c) CurvSegFlow, (d) U-Net++, and (e) nnU-Net, where true positives are shown in yellow, false positives in red, and false negatives in green.}
    \label{fig:seg_hard}
\end{figure*}

On the MicSim\_FluoMT-Complex dataset, where microtubule extremity intensities are often confounded with background fluorescence, CurvSegFlow substantially outperformed all other models, including nnU-Net (Table~\ref{tab:complex}). It achieved the highest Dice, MCC, and AUC-PR scores, along with superior precision. Although some models exhibited higher sensitivity, their precision was compromised, indicating a tendency to over-predict. For instance, ResUNet attained the highest sensitivity but showed considerably lower precision, reflecting over-segmentation. In contrast, CurvSegFlow effectively segmented dense and overlapping filaments while minimizing false positives (Figure ~\ref{fig:seg_hard}). This improvement was also reflected in topology-sensitive metrics, with CurvSegFlow achieving strong clDice and low HD95, indicating better preservation of filament continuity and reduced structural fragmentation compared to competing methods. While nnU-Net performed well, it remained slightly behind CurvSegFlow, particularly in precision, sensitivity, HD95 and AUC-PR, indicating that CurvSegFlow better prevents false negatives. Overall, CurvSegFlow demonstrated superior efficiency in segmenting microtubules in noisy images, highlighting its potential for reliable microtubule segmentation in challenging imaging conditions. 

\begin{table*}[H]
\centering
\footnotesize
\setlength{\tabcolsep}{3pt}  
\caption{Performance on the MicSim\_FluoMT-Complex dataset (mean across test images)}
\label{tab:complex}
\begin{tabular}{lccccccccccc}
\toprule
Model & Dice & IoU & SE & Prec. & SP & Acc. & MCC & AUC-ROC & AUC-PR & clDice & HD95\\
\midrule
U-Net 2015 \citep{ronneberger2015u} & 0.7681 & 0.6235 & 0.8953 & 0.6725 & 0.9824 & 0.9855 & 0.7658 & 0.9933 & 0.8939 & 0.8004 & 3.140 \\
U-Net++ 2018 \citep{zhou2018unet} & 0.7778 & 0.6377 & 0.8936 & 0.6887 & 0.9837 & 0.9860 & 0.7748 & 0.9937 & 0.8998 & 0.8089 & 3.110  \\
ResUnet 2018 \citep{li2019residual} & 0.7672 & 0.6224 & \textbf{0.9067} & 0.6649 & 0.9816 & 0.9856 & 0.7663 & 0.9939 & 0.8974 &0.8043 & 2.980  \\
CAR-Unet 2021 \citep{guo2021channel} & 0.6278 & 0.4600 & 0.7532 & 0.5621 & 0.9739 & 0.9651 & 0.6261 & 0.9723 & 0.7281 & 0.6540 & 15.340  \\
TransUNet 2021 \citep{chen2021transunet} & 0.6389 & 0.4691 & 0.8528 & 0.5107 & 0.9670 & 0.9746 & 0.6432 & 0.9841 & 0.7258 & 0.6656 & 5.096  \\
nnU-Net 2021 \citep{isensee2021nnu} & 0.8203 & 0.6952 & 0.6950 & 0.8604 & 0.9949 & 0.9867 & 0.8144 & 0.8894 & 0.8263 & 0.8227 & 2.439  \\
MedSegDiff 2024 \citep{wu2024medsegdiff} & 0.6991 & 0.5376 & 0.6979 & 0.7009 & 0.9880 & 0.9767 & 0.6872 & 0.7442 & 0.4890 & 0.6489 & 3.954  \\
\textbf{CurvSegFlow (Ours)} 
& \textbf{0.8223}
& \textbf{0.6984}
& 0.7838
& \textbf{0.8646}
& \textbf{0.9951}
& \textbf{0.9869}
& \textbf{0.8164}
& \textbf{0.9943}
& \textbf{0.9088}
& \textbf{0.8239}
& \textbf{2.235} \\
\bottomrule
\end{tabular}
\end{table*}

\begin{figure}
    \centering
    \includegraphics[width=0.99\linewidth]{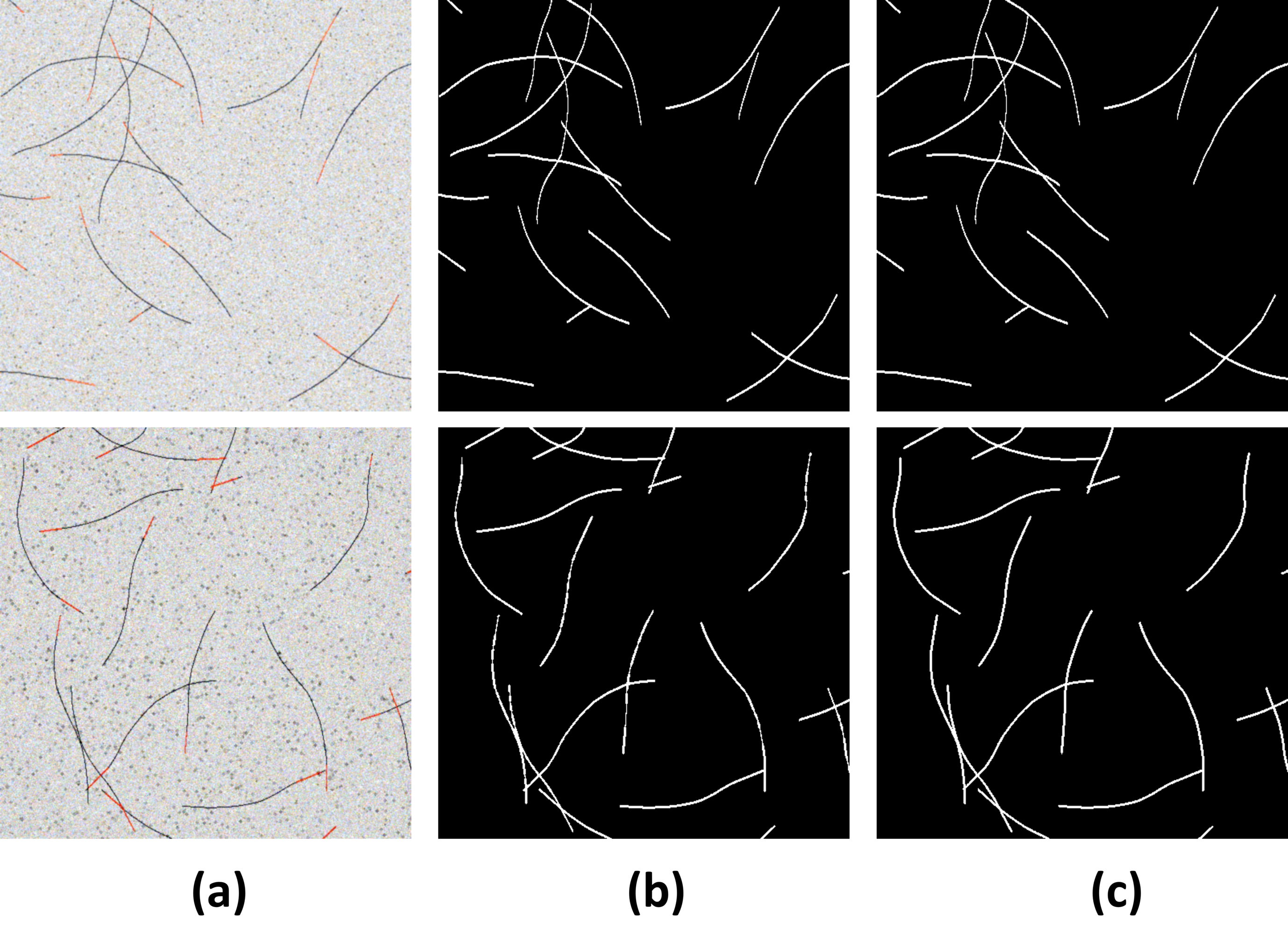}
    \caption{Segmentation on the SynthMT dataset: (a) Test image, (b) Ground truth, and (c) the predictions from CurvSegFlow.}
    \label{fig:seg_synthmt}
\end{figure}

We next investigated whether CurvSegFlow would remain competitive under high SNR conditions when trained on a reduced dataset. On the SynthMT dataset, where background noise is low, and using only 100 images for training, CurvSegFlow’s predictions closely matched the ground-truth masks (Figure \ref{fig:seg_synthmt}). The quantitative results confirmed its strong performance: Dice = 0.9604, IoU = 0.9250, sensitivity = 0.9641, precision = 0.9577, specificity = 0.9990, accuracy = 0.9981, MCC = 0.9597, AUC-ROC = 0.9965, and AUC-PR = 0.9808. Despite being trained on only 100 images, CurvSegFlow outperformed the foundation models reported in the SynthMT study \citep{koddenbrock2026synthetic}, including the best-performing model, SAM3Text+HPO, wich was initially trained on 5280 images followed by hyperparameter optimization on 10 synthetic SynthMT images  (average precision of 0.95). This validated CurvSegFlow’s excellent performance in high-SNR conditions with limited data.

\begin{figure*}
    \centering
    \includegraphics[width=0.99\linewidth]{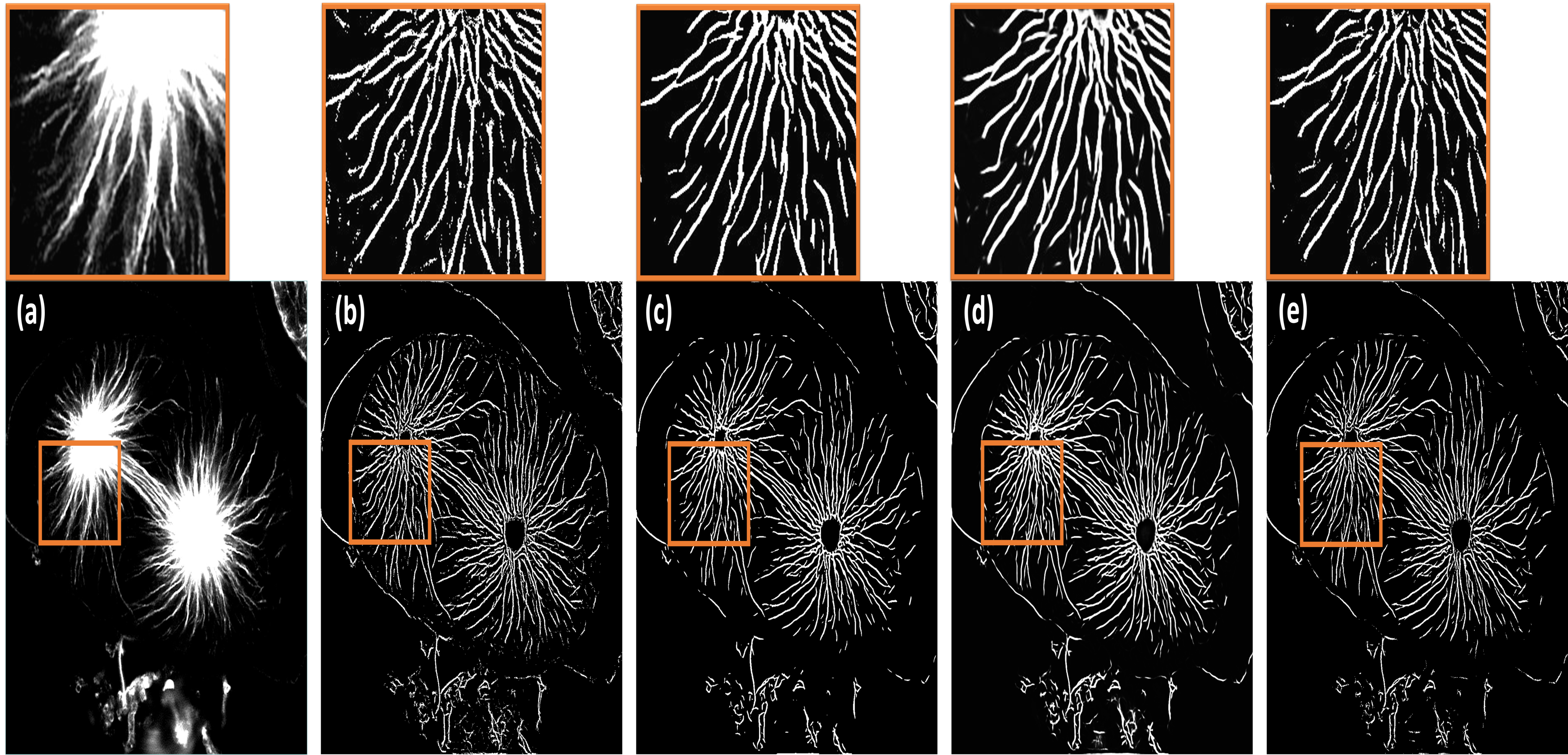}
    \caption{Segmentation on the MicReal\_FluoMT dataset: (a) Test image, (b) Ground truth, and (c-e) predictions from (c) CurvSegFlow, (d) U-Net++, and (e) nnU-Net.}
    \label{fig:seg_real}
\end{figure*}

We then challenged CurvSegFlow on real microscopy images using various datasets: MicReal\_FluoMT, IRM\_InVitro MT, and the revised Higaki 2024 datasets.  
MicReal\_FluoMT is demanding due to strong variability in fluorescence intensity across microtubules. Despite this, CurvSegFlow produced clean and consistent segmentations that often appeared less noisy than the annotations (Figure ~\ref{fig:seg_real}). In some regions, missing annotations caused correct predictions to be counted as false positives, which would lowered the measured precision.
Quantitatively, CurvSegFlow achieved the highest Dice, IoU, and sensitivity (Table~\ref{tab:micreal}). Although nnU-Net reported higher precision and AUC-ROC, CurvSegFlow provided the best trade-off between precision and sensitivity, detecting more true microtubules while maintaining good precision. This balance is important for this task, where missing thin or low-intensity filaments is more detrimental than producing a small number of extra detections, notably since these additional detections might be missing annotations.
CurvSegFlow also achieved a competitive MCC compared to nnU-Net and U-Net++, and outperformed U-Net and SynSeg, showing stable overall performance across both positive and negative classes. Its AUC-PR remained close to that of nnU-Net’s and superior to those of other models, confirming that it maintained strong precision across different recall levels.
As shown in Figure ~\ref{fig:seg_real}, CurvSegFlow produced more continuous filaments with fewer breaks and less fragmentation. This visual improvement was consistent with higher clDice scores and competitive HD95 values, supporting improved topological consistency in real microscopy conditions. These results indicated that CurvSegFlow is also well-suited for segmenting microtubules in real images, despite being trained on only 29 images. Furthermore, this work highlights that, in addition to handling background noise effectively (as seen with the MicSim\_FluoMT dataset), CurvSegFlow also copes well with low-contrast microtubules, making it a versatile tool.

\begin{table*}
\centering
\footnotesize
\setlength{\tabcolsep}{3pt}  
\caption{Performance comparison on the MicReal\_FluoMT dataset (mean across test images)}
\label{tab:micreal}
\begin{tabular}{lccccccccccc}
\toprule
Model & Dice & IoU & SE & Prec. & SP & Acc. & MCC & AUC-ROC & AUC-PR & clDice & HD95\\
\midrule
U-Net 2015 \citep{ronneberger2015u}    & 0.6617 & 0.4944 & 0.6434 & 0.6810 & 0.9704 & 0.9411 & 0.6297 & 0.9508 & 0.7165 & 0.6889 & 13.32\\
U-Net++ 2018 \citep{zhou2018unet}  & 0.7038 & 0.5430 & 0.6886 & 0.7197 & 0.9736 & 0.9481 & 0.6755 & 0.9473 & 0.7444 & 0.6931 & 12.17 \\
nnU-Net 2021 \citep{isensee2021nnu}   & 0.7040 & 0.5447 & 0.6496 & \textbf{0.7722} & \textbf{0.9811} & 0.9507 & 0.6813 & 0.9666 & \textbf{0.7943} & 0.7017 & \textbf{7.53} \\
Synseg 2025 \citep{guo2025synseg} & 0.5292 & 0.3632 & 0.4722 & 0.6240 & 0.9723 & 0.9262 & 0.5012 & 0.9247 & 0.5537 & - & - \\
\textbf{CurvSegFlow (Ours)} & \textbf{0.7093} & \textbf{0.5510} & \textbf{0.7176} & 0.7030 & 0.9745 & \textbf{0.9547} & \textbf{0.6853} & \textbf{0.9708} & 0.7826 & \textbf{0.7188} & 8.05 \\
\bottomrule
\end{tabular}%
\end{table*}

\begin{figure*}
    \centering
    \includegraphics[width=0.9\linewidth]{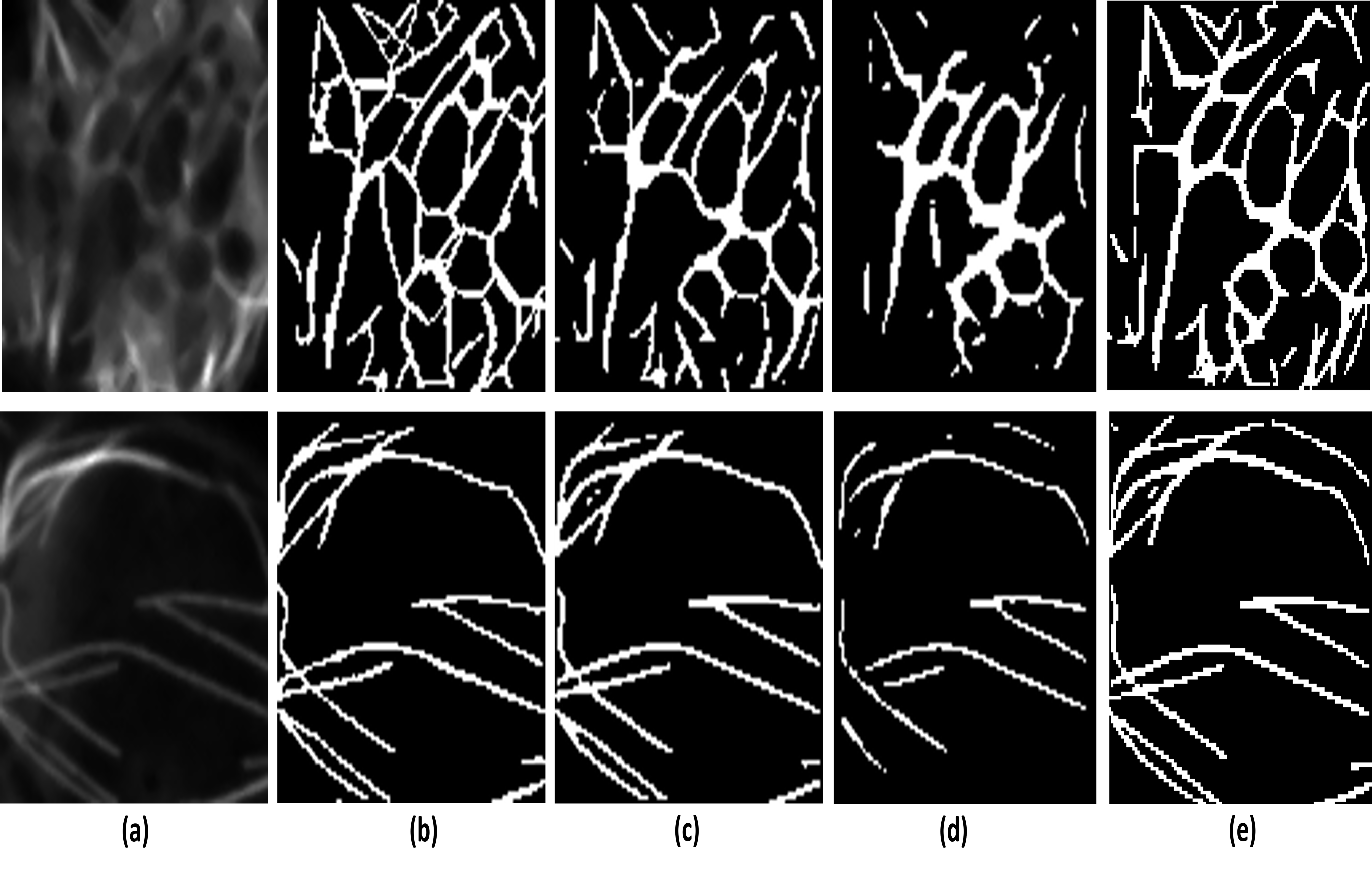}
    \caption{Segmentation on the revised Higaki 2024 dataset: (a) Test image, (b) Ground truth, and (c-e) predictions from (c) CurvSegFlow, (d) Synseg, and (e) nnU-Net}
    \label{fig:seg_higaki}
\end{figure*}

To further evaluate CurvSegFlow on real images, we used the revised Higaki 2024 dataset and performed five-fold cross-validation. CurvSegFlow achieved strong and consistent performance across folds, with the following metrics: Dice = 0.7435, AUC-PR = 0.8246, precision = 0.7561, sensitivity = 0.7393, clDice = 0.8135, and HD95 = 2.54. Specificity remained high at 0.9525, confirming effective minimization of false positives. As shown in Figure \ref{fig:seg_higaki}, on this dataset, CurvSegFlow outperformed SynSeg, a recently proposed model to segment cytoskeleton structures, which achieved an averaged Dice of $\sim$0.7 and precision of $\sim$0.65 \citep{guo2025synseg}. CurvSegFlow also slightly outperformed nnU-Net, particularly in precision, with the following scores: Dice = 0.7406, AUC-PR = 0.8244, precision = 0.7514, sensitivity = 0.7405, clDice = 0.8008, and HD95 = 2.18. These results demonstrated CurvSegFlow’s competitiveness in segmenting real microtubules with diverse filament networks and varying microscopy conditions.

\begin{figure*}
    \centering
    \includegraphics[width=0.99\linewidth]{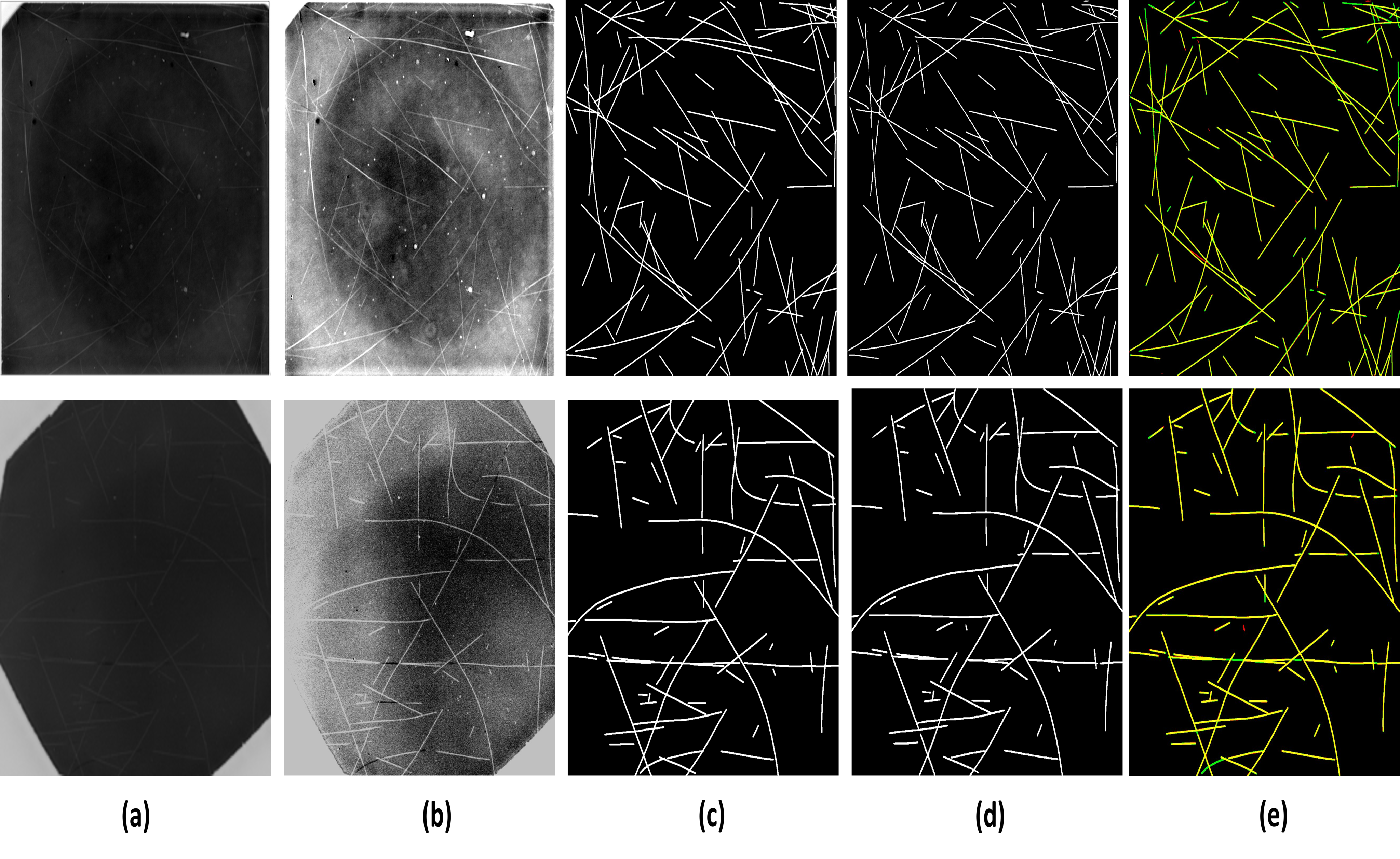}
    \caption{Segmentation on the IRM\_InVitroMT dataset: (a,b) Test images: (a) raw, (b) Contrast-enhanced, (c) Ground truths, (d) predictions from CurvSegFlow, and (e) composites, where true positives are shown in yellow, false positives in red, and false negatives in green.}
    \label{fig:seg_invitro}
\end{figure*}

Finally, we tested CurvSegFlow on a novel dataset, IRM\_In VitroMT, where microtubules are detected by light wave interference rather than fluorescent labeling. CurvSegFlow achieved strong performance, with Dice = 0.8365, AUC-PR = 0.9013, precision = 0.8542, sensitivity = 0.8236, clDice = 0.9381, and HD95 = 5.82. Despite variability in image appearance due to differences in sample preparation and imaging conditions, segmentation quality remained high (Figure \ref{fig:seg_invitro}), demonstrating robustness to diverse acquisition protocols.

Overall, CurvSegFlow generalizes well to various types of real microscopy images for microtubule segmentation. It effectively handled noise, intensity variation, and imperfect annotations while maintaining accurate and structurally consistent segmentations.

\subsection{Segmentation of other curvilinear structures}

To assess CurvSegFlow’s capability beyond microtubules, we evaluated it on benchmark datasets for other biomedical curvilinear structures: retinal vessels, corneal nerve fibers, and coronary arteries. Retinal vessel segmentation is a well‑studied task with numerous recent deep learning approaches demonstrating strong performance on DRIVE and CHASE\_DB1. Contemporary vessel models, including attention‑enhanced U‑Nets, hybrid convolution–transformer architectures and diffusion models, consistently reported AUC‑ROC values in the high 0.97 range on DRIVE and CHASE\_DB1, often prioritizing sensitivity to capture fine vessels \citep{ma2025vasca,hu2026multi,zhu2025automated}.  

\begin{figure*}[H]
    \centering
    \includegraphics[width=0.99\linewidth]{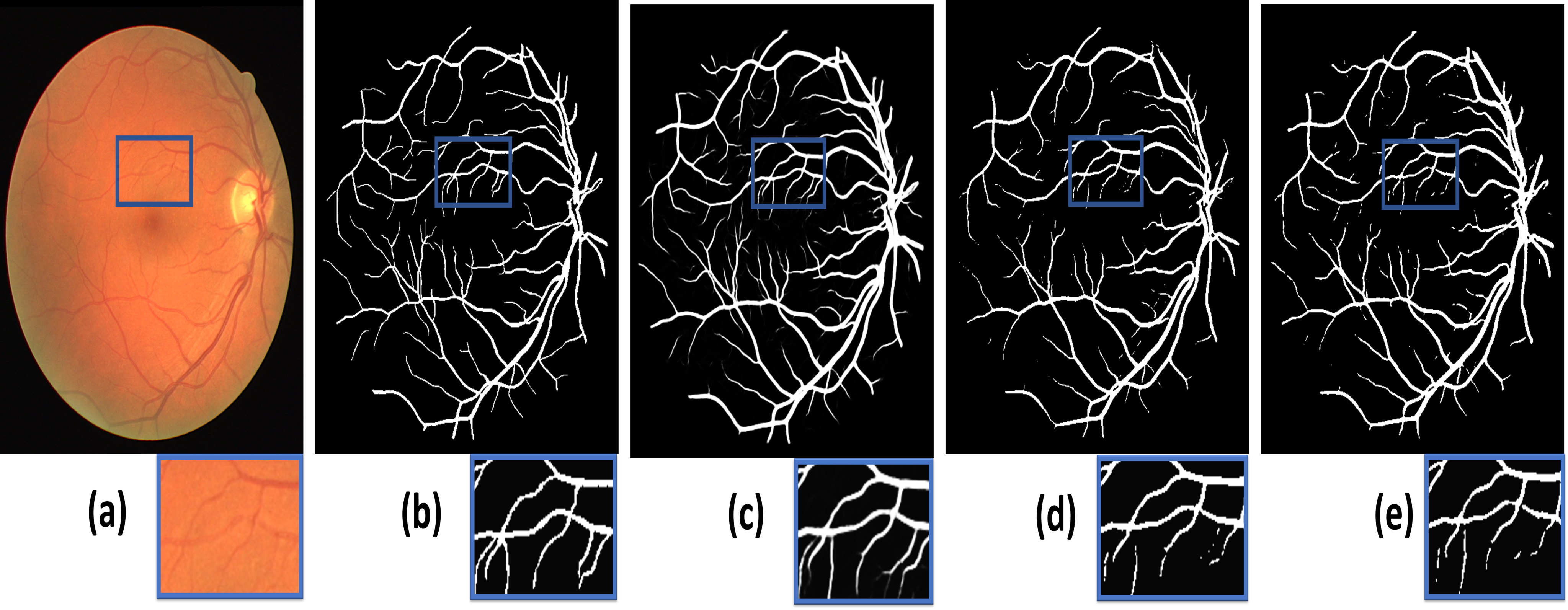}
    \caption{Segmentation on the DRIVE dataset: (a) Test image, (b) Ground truth, and (c-e) predictions from (c) CurvSegFlow, (d) CAR-Unet, and (e) U-Net.}
    \label{fig:seg_drive}
\end{figure*}

\begin{figure*}
    \centering
    \includegraphics[width=0.99\linewidth]{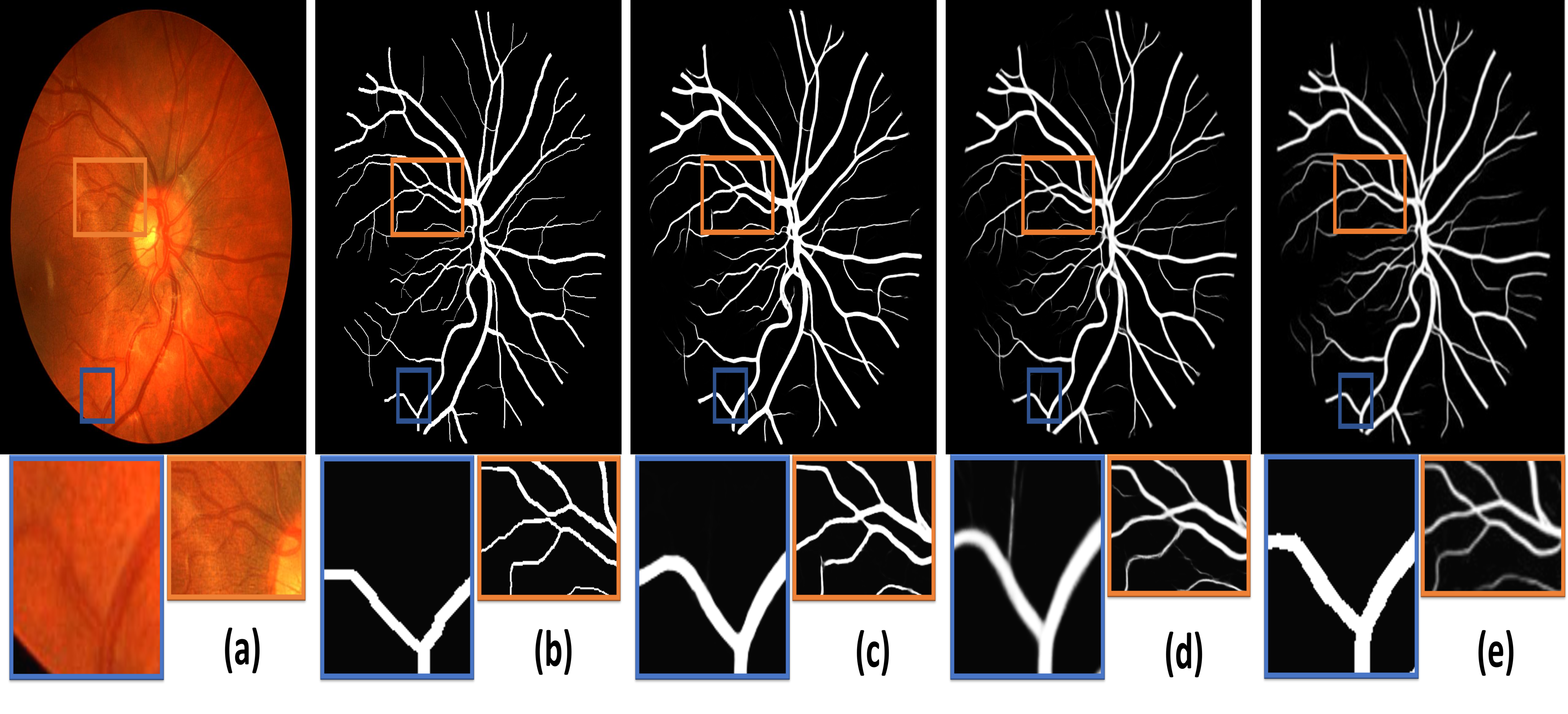}
    \caption{Segmentation on the CHASE\_DB1 dataset: (a) Test image, (b) Ground truth, and (c-e) predictions from (c) CurvSegFlow, (d) CAR-Unet, and (e) ARU-net.}
    \label{fig:seg_chasedb1}
\end{figure*}

On DRIVE, CurvSegFlow achieved the highest AUC‑ROC and accuracy, as well as the top sensitivity among all evaluated models (Table~\ref{tab:drive}), demonstrating competitive overall performance in a domain different from microtubules. Figure \ref{fig:seg_drive} highlights CurvSegFlow’s performance compared to previous published test predictions from U-Net and CAR-Unet. This comparison revealed that CurvSegFlow enables vessel continuity and edge preservation, key focuses of recent models such as in \citep{hu2026multi}. Likewise, on CHASE\_DB1, CurvSegFlow attained the highest Dice, IoU, sensitivity, and AUC-ROC scores (Table~\ref{tab:chasedb1}), confirming that its strong performance persisted despite changes in  dataset characteristics and vessel distributions, as illustrated in Figure \ref{fig:seg_chasedb1}. Compared to recent retinal vessel segmentation works, which often report minor variations in Dice and AUC-ROC due to architectural enhancements such as plenary attention or multi‑scale fusion, CurvSegFlow’s performance aligned well with these state‑of‑the‑art results without requiring dataset‑specific design choices.  

\begin{table*}
\centering
\footnotesize
\setlength{\tabcolsep}{3pt}  
\caption{Performance on the DRIVE dataset}
\label{tab:drive}
\begin{tabular}{lcccccccccc}
\toprule
Model & Dice & IoU & SE & Prec. & SP & Acc. & MCC & AUC-ROC & AUC-PR \\
\midrule
U-Net 2015 \citep{ronneberger2015u} & 0.8021 & 0.6699 & 0.8247 & 0.7806 & 0.9788 & 0.9659 & 0.7838 & 0.9842 & 0.8888 \\
Wave-Net 2016 \citep{liu2023wave} & 0.8254 & 0.703 & 0.8164 & 0.737 & 0.9764 & 0.9561 & 0.761 & 0.975 & -- \\
U-Net++ 2018 \citep{zhou2018unet}& 0.8107 & 0.6769 & 0.8119 & 0.8096 & 0.9825 & 0.9682 & 0.7934 & 0.9863 & 0.8989 \\
AttU-Net 2018 \citep{oktay2018attention} & 0.8116 & 0.6827 & 0.7809 & 0.8448 & 0.9809 & 0.9548 & 0.7948 & 0.9782 & 0.9072 \\
R2U-Net 2018\citep{alom2018recurrent} & 0.8171 & 0.6905 & 0.7792 & 0.8589 & 0.9813 & 0.9556 & 0.7982 & 0.9784 & - \\
ResU-Net 2019 \citep{li2019residual} & 0.8237 & 0.7000 & 0.7969 & 0.8524 & 0.9799 & - & 0.8027 & 0.9799 & - \\
Wu et al. 2019 \citep{wang2019dual} & 0.8270 & 0.705 & 0.7940 & 0.8629 & 0.9816 & 0.9567 & 0.8044 & 0.9772 & - \\
IterNet 2020 \citep{li2020iternet} & 0.8218 & 0.6975 & 0.7791 & 0.8695 & 0.9831 & 0.9574 & 0.8037 & 0.9813 & - \\
CS\textsuperscript{2}Net 2020 \citep{mou2021cs2} & 0.8070 & 0.675 & 0.8218 & 0.792 & \textbf{0.9890} & 0.9632 & 0.773 & 0.9825 & - \\
Swin-UNet 2021 \citep{cao2022swin} & 0.8210 & 0.6967 & 0.8005 & 0.8426 & 0.9881 & 0.9543 & 0.8032 & 0.9778 & - \\
TransUNet 2021 \citep{chen2021transunet} & 0.8227 & 0.6989 & 0.7992 & 0.8476 & 0.9790 & 0.9561 & 0.8034 & 0.9797 & - \\
CSU-Net 2021 \citep{wang2020csu} & 0.8251 & 0.7022 & 0.8071 & 0.8439 & 0.9782 & 0.9565 & 0.8013 & 0.9801 & - \\
MD-Net 2021 \citep{shi2021md} & 0.8199 & 0.6948 & 0.7730 &  \textbf{0.8729} & 0.8727 & 0.9568 & 0.6443 & 0.9807 & - \\
ARU-net 2022 \citep{wu2022atrous}
  & 0.8179 & 0.6919 & 0.8043 & 0.8324 & 0.9844 & 0.9686 & 0.8009 & 0.9842 & -- \\
AFNet 2023 \citep{li2023retinal} & 0.7990 & 0.669 & 0.8139 & 0.784 & 0.9818 & 0.9580 & 0.764 & 0.9820 & - \\
ResDO-UNet 2023 \citep{liu2023resdo} & 0.8229 & 0.6991 & 0.7985 & 0.8488 & 0.9791 & 0.9561 & 0.8037 & - & - \\
IterMiU-Net 2023\citep{kumar2023itermiunet} & 0.8277 & 0.7059 & 0.8006 & 0.8567 & 0.9805 & 0.9575 & \textbf{0.8069} & 0.9811 & - \\
YoloCurvSeg 2023 \citep{lin2023yolocurvseg} & 0.8184 & 0.6926 & 0.8199 & 0.8169 & 0.9790 & 0.9627 & 0.7976 & - & - \\
RCAR-UNet 2024 \citep{ding2024rcar} & 0.8047 & 0.6734 & 0.7487 & 0.8698 & 0.9836 & 0.9537 & 0.7859 & - & - \\
SemFlow 2024 \citep{wang2024semflow} & 0.454 & 0.295 & - & - & - & - & - & - & - \\
VasCA-Net 2025 \citep{ma2025vasca} & \textbf{0.8301} & \textbf{0.7094} & 0.8145 & 0.8463 & 0.9784 & 0.9576 & 0.8068 & 0.9815 & \textbf{0.9173} \\
FlowSDF 2025 \citep{bogensperger2025flowsdf} & 0.776 & 0.635 & - & - & - & - & - & - & - \\
FMS$^2$ 2026 \citep{asadi2026fms} & 0.789 & 0.653 & - & - & - & - & - & - & - \\

\textbf{CurvSegFlow (Ours)} & 0.8174 & 0.6921 & \textbf{0.8317} & 0.8036 & 0.9814 & \textbf{0.9689} & 0.8005 & \textbf{0.9872} & 0.9016 \\

\bottomrule
\end{tabular}
\end{table*}

\begin{table*}
\centering
\footnotesize
\setlength{\tabcolsep}{3pt}
\caption{Performance on the CHASE\_DB1 dataset}
\label{tab:chasedb1}
\begin{tabular}{lcccccccccc}
\toprule
Model & Dice & IoU & SE & Prec. & SP & Acc. & MCC & AUC-ROC & AUC-PR \\
\midrule
U-Net 2015 \citep{ronneberger2015u}
  & 0.7783 & 0.6371 & 0.8288 & 0.7335 & 0.8288 & 0.9578 & 0.5173 & 0.9578 & -- \\
LadderNet 2018 \citep{zhuang2018laddernet}
  & 0.8159 & 0.6891 & 0.7640 & 0.8752 & 0.9865 & 0.9620 & 0.7967 & 0.9832 & -- \\
AttU-Net 2018 \citep{oktay2018attention}
  & 0.8147 & 0.6873 & 0.7591 & 0.8791 & 0.9881 & 0.9629 & 0.7966 & 0.9848 & -- \\
R2U-Net 2018 \citep{alom2018recurrent}
  & 0.7928 & 0.6567 & 0.7756 & 0.8108 & 0.9820 & 0.9634 & 0.7840 & 0.9815 & -- \\
Dense-UNet 2019 \citep{qiang2019k}
  & 0.8174 & 0.6912 & 0.7707 & 0.8703 & 0.9857 & 0.9620 & 0.7965 & 0.9820 & -- \\
Cherukuri et al. 2020 \citep{cherukuri2019deep}
  & 0.8211 & 0.6965 & 0.8025 & 0.8407 & 0.9874 & 0.9693 & 0.8207 & 0.9658 & -- \\
CTF-Net 2020 \citep{wang2020ctf}
  & 0.8220 & 0.6977 & 0.7948 & 0.8511 & 0.9842 & 0.9648 & 0.8041 & 0.9847 & -- \\
IterNet 2020 \citep{li2020iternet}
  & 0.8073 & 0.6769 & 0.7970 & 0.8179 & 0.9823 & 0.9655 & 0.7993 & 0.9851 & -- \\
D-GaussianNet 2021 \citep{alvarado2021d}
  & 0.8077 & 0.6773 & 0.7530 & 0.8710 & 0.9863 & \textbf{0.9798} & 0.7872 & -- & -- \\
TransUNet 2021 \citep{chen2021transunet}
  & 0.8192 & 0.6938 & 0.7459 & \textbf{0.9087} & \textbf{0.9903} & 0.9637 & 0.7959 & 0.9852 & -- \\

CAR-Unet 2021 \citep{guo2021channel}
  & 0.8099 & 0.6805 & 0.8439 & 0.7786 & 0.9839 & 0.9751 & 0.7974 & 0.9898 & -- \\
  
Swin-UNet 2022 \citep{cao2022swin}
  & 0.8201 & 0.6950 & 0.8241 & 0.8161 & 0.9879 & 0.9721 & \textbf{0.8390} & 0.9832 & -- \\
  
ARU-net 2022 \citep{wu2022atrous}
  & 0.8005 & 0.6676 & 0.8099 & 0.7916 & 0.9856 & 0.9746 & 0.7862 & 0.9869 & -- \\
  
ResDO-UNet 2023 \citep{liu2023resdo}
  & 0.8236 & 0.6999 & 0.8020 & 0.8464 & 0.9794 & 0.9672 & 0.7940 & -- & -- \\
DDPM 2023 \citep{alimanov2023denoising}
  & 0.7384 & 0.5889 & 0.7007 & 0.7803 & -- & 0.9649 & -- & -- & -- \\
MDUNet 2023 \citep{jayachandran2023multi}
  & 0.6388 & 0.4693 & 0.8111 & 0.5269 & 0.9817 & 0.9660 & 0.8089 & -- & -- \\
IterMiU-Net 2023 \citep{kumar2023itermiunet}
  & 0.8263 & 0.7040 & 0.8134 & 0.8397 & 0.9820 & 0.9646 & 0.8101 & 0.9852 & -- \\
RCAR-UNet 2024 \citep{ding2024rcar}
  & 0.7470 & 0.5963 & 0.7475 & 0.7465 & 0.9798 & 0.9566 & 0.7591 & -- & -- \\
RVS-FDSC 2024 \citep{kong2024rvs}
  & 0.8050 & 0.6736 & 0.8356 & 0.7767 & 0.9856 & 0.9743 & 0.8334 & 0.9867 & -- \\
WS-DMF 2024 \citep{tan2024deep}
  & -- & -- & 0.7841 & -- & 0.9797 & 0.9566 & 0.7884 & -- & -- \\
GViTRSNet 2025 \citep{li2025gvit}
  & 0.7981 & 0.6641 & -- & -- & -- & 0.8386 & -- & -- & -- \\
GEA-UNet 2025 \citep{roy2024new}
  & 0.7578 & 0.6156 & 0.8403 & 0.6901 & 0.9814 & 0.9696 & 0.8237 & -- & -- \\
MSTP-Net 2025 \citep{wang2025multi}
  & 0.8074 & 0.6771 & 0.8485 & 0.7701 & 0.9830 & 0.9745 & 0.8373 & -- & -- \\
Zhu et al. 2025 \citep{zhu2025automated}
  & 0.8002 & 0.6669 & 0.8099 & 0.7909 & 0.9857 & 0.9744 & 0.7956 & 0.9865 & -- \\
MVM-UNet (2026) \citep{hu2026multi}
  & 0.8146 & 0.6872 & 0.8305 & 0.7993 & 0.9860 & 0.9762 & 0.8388 & -- & -- \\

\textbf{CurvSegFlow (Ours)}
  & \textbf{0.8429} & \textbf{0.7285} & \textbf{0.8652} & 0.8218 & 0.9857 & 0.9772 & 0.8309 & \textbf{0.9921} & \textbf{0.9215} \\

\end{tabular}
\end{table*}

\begin{figure*}
    \centering
    \includegraphics[width=0.8\linewidth]{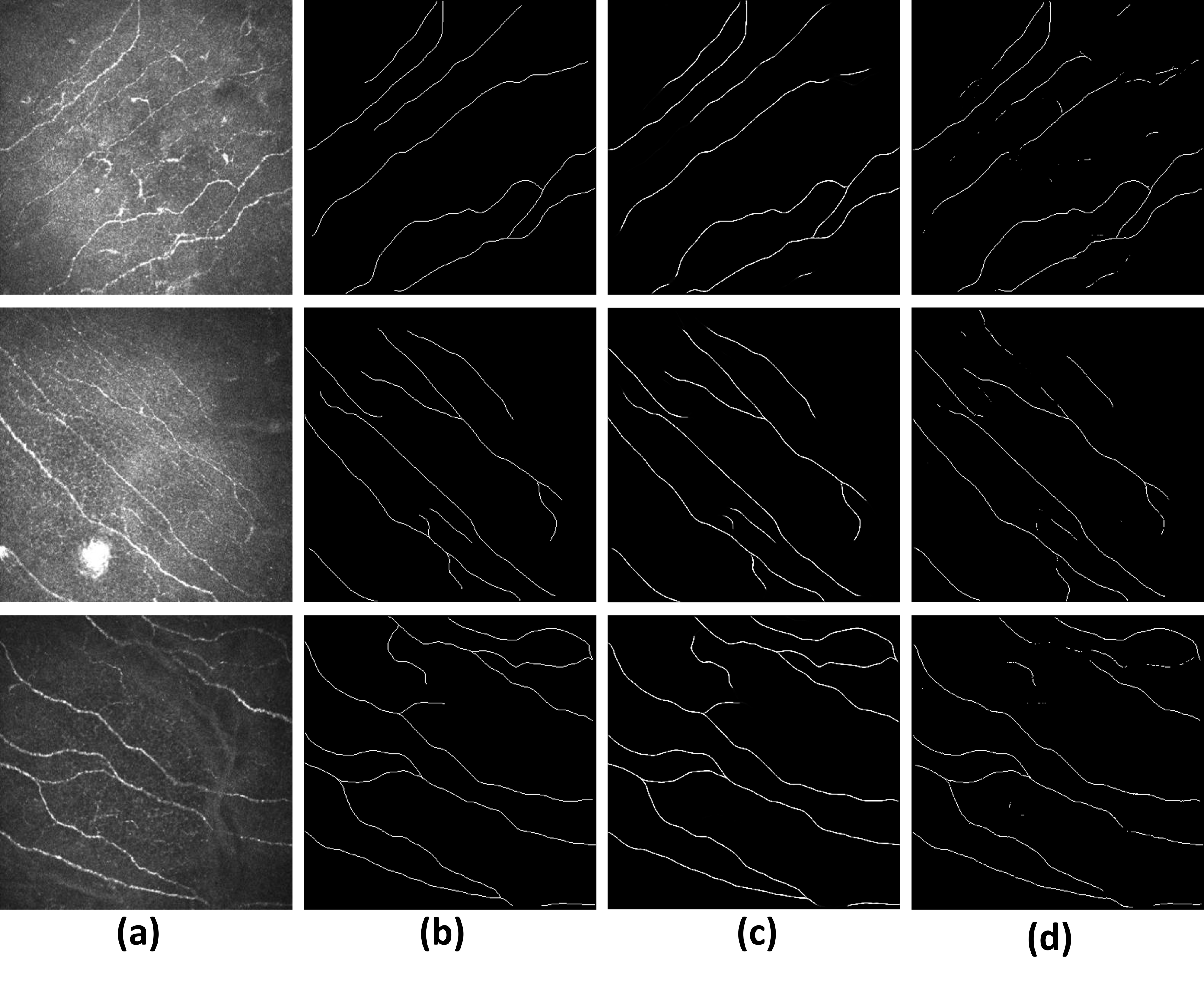}
    \caption{Segmentation on the CORN1 dataset: (a) Test image, (b) Ground truth, and (c-d) predictions from (c) CurvSegFlow, and (d) U-Net.}
    \label{fig:seg_corn1}
\end{figure*}

For corneal nerve fibers, publicly available benchmarks like CORN‑1 provide a less mature standard but still highlight the challenges of segmenting extremely thin, tortuous structures in noisy backgroud with limited labeled data. Deep learning methods developed specifically for nerve fiber segmentation typically rely on recent U‑Net variants and focus on preserving connectivity and morphological features \citep{oktay2018attention, wang2022uctransnet}. On CORN‑1, CurvSegFlow substantially outperformed all other models across most evaluation metrics, including Dice, sensitivity, MCC, and AUC‑ROC (Table~\ref{tab:corn1}). This indicated that CurvSegFlow captures nerve morphology effectively, as illustrated in Figure \ref{fig:seg_corn1}, in a context where existing methods, even specialized ones, struggle with class imbalance and fine structural continuity in noisy images \citep{liu2025lightweight}.

\begin{table*}
\centering
\footnotesize
\setlength{\tabcolsep}{3pt}  
\caption{Performance on the CORN-1 dataset}
\label{tab:corn1}
\begin{tabular}{lcccccccccc}
\toprule
Model & Dice & IoU & SE & Prec. & SP & Acc. & MCC & AUC-ROC & AUC-PR \\
\midrule
U-Net 2015 \citep{ronneberger2015u} & 0.5260 & 0.3613 & 0.5484 & 0.5054 & 0.9852 & 0.9781 & 0.5140 & 0.7287 & - \\ 
M-Net 2017 \citep{mehta2017m} & 0.5110 & 0.3482 & 0.5310 & 0.4925 & 0.9831 & 0.9760 & 0.4985 & 0.7002 & - \\ 
Pix2Pix cGAN 2017 \citep{isola2017image} & 0.5419 & 0.3750 & 0.5620 & 0.5232 & 0.9902 & 0.9835 & 0.5320 & 0.7580 & - \\
Att-UNet 2018 \citep{oktay2018attention} & 0.5165 & 0.3530 & 0.5350 & 0.4992 & 0.9845 & 0.9772 & 0.5052 & 0.7100 & - \\  
CNS-Net 2020 \citep{wei2020deep} & 0.4420 & 0.2894 & 0.4602 & 0.4252 & 0.9805 & 0.9723 & 0.4304 & 0.6810 & - \\ 
CS\textsuperscript{2}Net 2020 \citep{mou2021cs2} & -- & -- & 0.8398 & \textbf{0.7446} & -- & -- & -- & -- & - \\ 
DAANet 2022 \citep{liang2022fusion} & 0.5125 & 0.3494 & 0.5285 & 0.4975 & 0.9862 & 0.9789 & 0.5015 & 0.7156 & - \\ 
UCTransNet 2022 \citep{wang2022uctransnet} & 0.4161 & 0.2680 & 0.4350 & 0.3988 & 0.9754 & 0.9670 & 0.4022 & 0.6476 & - \\ 
YoloCurvSeg 2023 \citep{lin2023yolocurvseg} & 0.6719 & 0.5059 & 0.6760 & 0.6678 & 0.9844 & 0.9798 & 0.6610 & 0.8302 & - \\ 
PCT-Net 2025 \citep{liu2025lightweight} & 0.5656 & 0.3987 & 0.5890 & 0.5440 & 0.9915 & 0.9851 & 0.5552 & 0.7454 & - \\ 

\textbf{CurvSegFlow (Ours)} & \textbf{0.7600} & \textbf{0.6129} & \textbf{0.8652} & 0.6776 & \textbf{0.9946} & \textbf{0.9930} & \textbf{0.7623} & \textbf{0.9969} & \textbf{0.8249} \\
\bottomrule
\end{tabular}
\end{table*}

\begin{figure}
    \centering
    \includegraphics[width=0.9\linewidth]{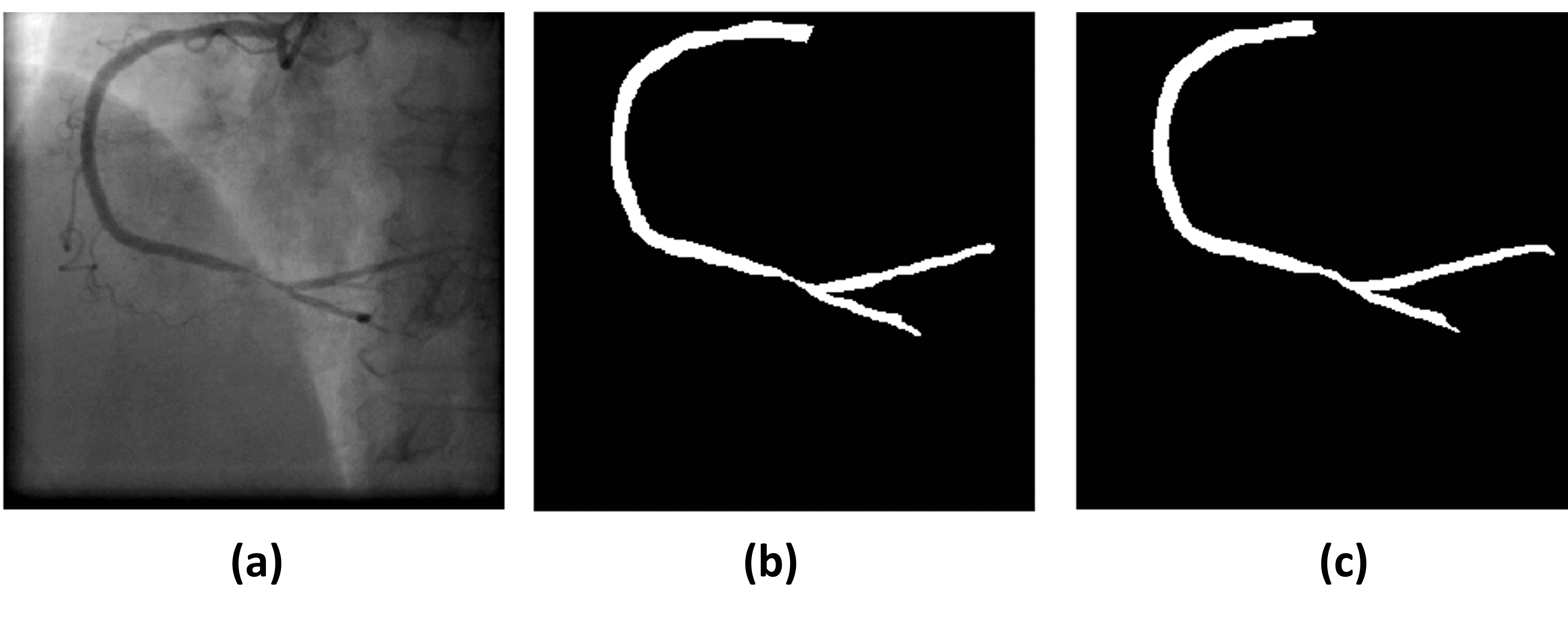}
    \caption{Segmentation on the ARCADE dataset: (a) Test image, (b) Ground truth, and (c) prediction from CurvSegFlow.}
    \label{fig:seg_arcade}
\end{figure}

Finally, we evaluated CurvSegFlow for coronary artery segmentation using the ARCADE dataset. CurvSegFlow achieved the best performance across most evaluation metrics (Table~\ref{tab:arcade}). Improvements were consistent in complementary measures, including sensitivity and precision, indicating accurate delineation of vessel structures with limited false positives. The high AUC-ROC and AUC-PR further suggested stable performance across decision thresholds.
These results demonstrated that CurvSegFlow effectively captures the geometric and topological characteristics of coronary vessels, as illustrated in Figure \ref{fig:seg_arcade}. This behavior may be attributed to its ability to model long-range dependencies while preserving local continuity, which is critical in this task. Notably, this performance is obtained without task-specific architectural adaptations, underscoring its robustness to complex vascular structures. 

\begin{table*}
\centering
\footnotesize
\setlength{\tabcolsep}{3pt}
\caption{Performance on the ARCADE dataset}
\label{tab:arcade}
\begin{tabular}{lccccccccc}
\toprule
Model & Dice & IoU & SE & Prec. & SP & Acc. & MCC & AUC-ROC & AUC-PR \\
\midrule
U-Net 2015 \citep{ronneberger2015u}         & 0.5523 & 0.5153 & 0.5325 & 0.5734 & 0.5226 & 0.9656 & 0.4101 & - & - \\
U-Net++ 2018 \citep{zhou2018unet}         & 0.7387 & 0.5856 & 0.7143 & 0.7647 & 0.7647 & 0.9808 & 0.6742 & - & - \\
TransUNet 2021 \citep{chen2021transunet}    & 0.5889 & 0.4364 & 0.5141 & 0.6925 & 0.7569 & 0.9762 & 0.5321 & - & - \\
TransFuse 2021 \citep{chen2021transunet}   & 0.7210 & 0.5737 & \textbf{0.8511} & 0.6247 & 0.9819 & 0.9769 & 0.7032 & - & - \\
nnU-Net 2021 \citep{isensee2021nnu}         & 0.7119 & 0.5527 & 0.6997 & 0.7247 & 0.9865 & 0.9713 & 0.6943 & - & - \\
MALUNet 2022 \citep{ruan2022malunet}        & 0.7147 & 0.5561 & 0.6801 & 0.7532 & 0.9915 & 0.9801 & 0.6993 & - & - \\
MISSFormer 2022 \citep{huang2022missformer} & 0.3680 & 0.2381 & 0.2704 & 0.5765 & 0.6863 & 0.9693 & 0.2985 & - & - \\
VM-UNet 2024 \citep{ruan2024vm}             & 0.7051 & 0.5445 & 0.6987 & 0.7117 & 0.9886 & 0.9773 & 0.6876 & - & - \\
SAM-VMNet 2025 \citep{huang2025deep}                            & 0.7733 & 0.6303 & 0.7343 & \textbf{0.8162} & \textbf{0.9933} & 0.9832 & 0.7421 & - & - \\
\textbf{CurvSegFlow (Ours)}             & \textbf{0.7737} & \textbf{0.6312} & 0.7675 & 0.7839 & 0.9917 & \textbf{0.9835} & \textbf{0.7652} & \textbf{0.9877} & \textbf{0.8353} \\
\bottomrule
\end{tabular}
\end{table*}

\vspace{0.2cm}

Overall,  CurvSegFlow’s performances in segmenting other curvilinear structures demonstrated its versatility and robustness, enabling accurate segmentations under various imaging conditions, noise and contrast levels, and across diverse geometries and network structures.

\subsection{Ablation Studies}
Previous works based on U-Net architectures for curvilinear structure segmentation using a single forward pass have shown that adding residual connections or attention mechanisms improves segmentation performance (e.g., CAR-UNet \citep{guo2021channel}  or CS\textsuperscript{2}Net \citep{mou2021cs2}). We investigated whether such modules would also benefit the dynamic learning of the vector field in CurvSegFlow. Table~\ref{tab:ablation_drive} presents an ablation study conducted on the DRIVE dataset, examining the contributions of residual connections and attention gates within the proposed flow-matching U-Net. The results showed that adding attention gates to the standard U-Net yielded the best overall performance, achieving the highest Dice, IoU, sensitivity, MCC, and AUC-PR scores. The only exception was precision, where U-Net + ResNet + Attention scored marginally higher (0.8073 vs.\ 0.8036), suggesting that residual connections can slightly improve false-positive suppression when combined with attention, albeit at the cost of reduced sensitivity. 
The effectiveness of attention gates aligns with the nature of curvilinear structures. In biomedical images such as retinal fundus photographs, corneal nerve maps, or fluorescence microscopy images of microtubules, the structures of interest are thin, elongated, and occupy a small fraction of the image. This creates a strong class imbalance between vessel and background pixels. The attention gates applied to the skip connections directly address this issue by selectively reweighting encoder features before they are passed to the decoder, allowing the network to suppress background activations and concentrate on vessel-relevant patterns. The decoder therefore receives more focused feature maps, which enhances the reconstruction of fine and continuous structures. 
Adding residual blocks provided consistent but modest improvements over the plain U-Net baseline across all metrics compared to attention gates, confirming that residual connections do not degrade performance. However, their benefit was limited in this context. Residual connections primarily help stabilize gradient flow in very deep networks, while the proposed U-Net is relatively shallow. More importantly, they operate within each convolutional block and do not directly filter the skip connections, so they do not address the core challenge of background suppression in class-imbalanced segmentation. When combined with attention gates, the two mechanisms partially conflict: residual shortcuts forward all features indiscriminately across layers, while attention gates aim to suppress irrelevant ones on the skip paths, potentially diluting the gating effect. Given that the number of parameters in U-Net with attention was 8.588 millions compared to 8.935 millions with attention and residual, we concluded that adding only attention gates in the decoder was the optimal configuration for this segmentation task. 

\begin{figure*}
    \centering
    \includegraphics[width=0.99\linewidth]{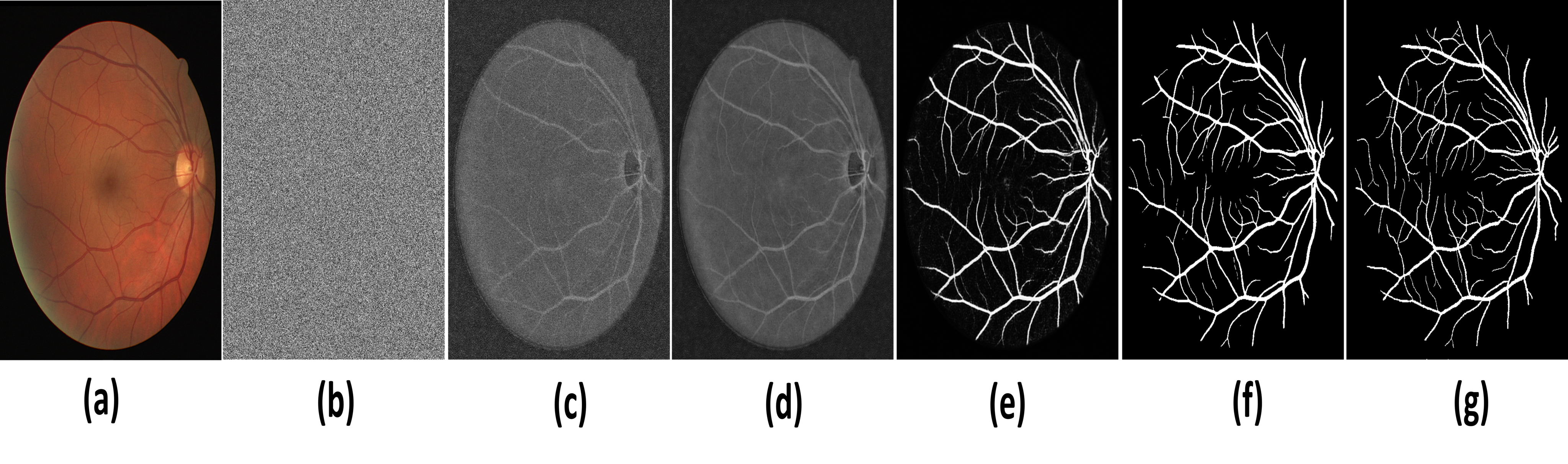}
    \caption{CurvSegFlow trajectory on a DRIVE image: (a) input, (b-f) evolution of the predicted segmentation over time, from an initial noisy state ($t$=0) to the final prediction ($t$=1), including intermediate steps at $t$= 0.25, 0.33, 0.67, and (g) ground truth.}
    \label{fig:flow_steps}
\end{figure*}

\begin{table*}[t]
\centering
\footnotesize
\caption{Ablation study of architectural components of CurvSegFlow on the DRIVE dataset. Best results are shown in bold.}
\label{tab:ablation_drive}

\begin{tabular}{lcc ccccccc}
\hline
Model & Residual & Attention & Dice & IoU & SE & Prec. & MCC & AUC-PR & Parameters (M)\\
\hline

U-Net & $\times$ & $\times$
& 0.8086 & 0.6786 & 0.8245 & 0.7933 & 0.7909 & 0.8925 & \textbf{8.501}\\

U-Net  
& $\checkmark$ & $\times$
& 0.8098 & 0.6801 & 0.8292 & 0.7912 & 0.7922 & 0.8949 & 8.848 \\

U-Net  & $\times$ & $\checkmark$
& \textbf{0.8174} & \textbf{0.6912} & \textbf{0.8317} & 0.8036 & \textbf{0.8005} & \textbf{0.9016} & 8.588 \\

U-Net & $\checkmark$ & $\checkmark$
& 0.8147 & 0.6873 & 0.8222 & \textbf{0.8073} & 0.7976 & 0.8990 & 8.935\\

\hline
\end{tabular}
\end{table*}

We next evaluated the benefits of implementing a triple-loss approach in CurvSegFlow, comprising a MSE loss for vector field estimation and a combination of weighted BCE and Dice losses for the reconstructed mask.
Table~\ref{tab:ablation_loss} displays the impact of different loss function combinations on the DRIVE dataset. Using MSE alone led to a collapse in segmentation performance, despite a high precision, indicating that the model predicted predominantly background pixels and failed to capture vessel structures. This suggests that supervising only the vector field was insufficient for this segmentation task.
Introducing Dice loss improved performance, as it directly optimizes region overlap but is insensitive to class imbalance. Similarly, combining MSE with weighted BCE achieved Dice and IoU scores comparable to those of MSE with Dice, while significantly improving precision and AUC-PR, reflecting better control over false positives. This suggested that weighted BCE effectively penalizes misclassification of minority vessel pixels.
The best overall performance was obtained by combining MSE, Dice, and weighted BCE, which attained the highest Dice, IoU, MCC, and AUC-PR. This combination leverages complementary properties: Dice enforces structural overlap, weighted BCE improves class balancing and pixel-wise discrimination, and MSE stabilizes regression of the flow field. Together, they provide a balanced optimization objective that improves both sensitivity to thin structures and robustness to background noise.

\begin{table*}[t]
\centering
\footnotesize
\caption{Ablation study of loss function combinaisons on the DRIVE dataset. Best results are shown in bold.}
\label{tab:ablation_loss}

\begin{tabular}{lccc cccccc}
\hline
\footnotesize
MSE & Dice loss & wBCE & Dice score  & IoU & SE & Prec. & MCC & AUC-PR \\
\hline

$\checkmark$ 
& $\times$ & $\times$
& 0.0194 & 0.0097 & 0.0098 & \textbf{0.8989} & 0.0889 & 0.6902 \\

$\checkmark$ & $\checkmark$ & $\times$
& 0.8091 & 0.6790 & \textbf{0.8363} & 0.7835 & 0.7915 & 0.8262 \\

$\checkmark$ & $\times$ & $\checkmark$
& 0.8082 & 0.6782 & 0.7976 & 0.8190 & 0.7910 & 0.8969 \\

$\checkmark$ & $\checkmark$ & $\checkmark$
& \textbf{0.8174} & \textbf{0.6912} & 0.8317 & 0.8036 & \textbf{0.8005} & \textbf{0.9017} \\

\hline
\end{tabular}
\end{table*}

We analyzed the impact of the number of integration steps during inference, as reported in Table~\ref{tab:results_ts}. CurvSegFlow is robust to the choice of time-step discretization, with relatively stable performance across all settings.
A small number of steps was sufficient to achieve strong segmentation quality, with a time step number equal to 3, leading to the best Dice, precision, MCC and AUC-PR. Increasing the number of iterations did not lead to consistent improvements; rather, it shifted the balance between sensitivity and precision toward over-segmentation.
We concluded that CurvSegFlow does not require many refinement steps to reach optimal performance, confirming the efficiency of the proposed dynamic formulation and its suitability for fast inference.

\begin{table}[t]
\centering
\scriptsize
\caption{Ablation study across different time steps (N) on the DRIVE dataset. Best results are shown in bold.}
\label{tab:results_ts}

\begin{tabular}{lcccccc}
\hline
N & Dice & SE & Pre & MCC & AUC-PR & Infer. time (s)\\
\hline

1  & 0.7975 & 0.8186 & 0.7774 & 0.7788 & 0.8798 & \textbf{0.0494}\\

3  & \textbf{0.8174} & 0.8317 & \textbf{0.8036} & \textbf{0.8005} & \textbf{0.9019} & 0.1207\\

5  & 0.8134 & 0.8241 & 0.8030 & 0.7962 & 0.8988 & 0.2183\\

10 & 0.8133 & \textbf{0.8494} & 0.7801 & 0.7963 & 0.8999 & 0.4648\\

\hline
\end{tabular}
\end{table}

Overall, these ablation studies demonstrated that attention mechanisms, composite loss functions, and time step number optimization address complementary challenges. While attention gates enhance feature selection and spatial focus within the network, the combined loss formulation ensures effective optimization under severe class imbalance. Their joint contribution was essential for achieving state-of-the-art performance in curvilinear structure segmentation.

\section{Conclusion and Perspectives}

In this work, we introduce CurvSegFlow, a flow-based framework for segmenting curvilinear structures in biomedical images. By formulating segmentation as a time-dependent reconstruction process, the proposed method progressively refines a noisy initialization into the target structure through a learned velocity field (Figure 15). This dynamic approach contrasts with conventional single-pass segmentation, enabling gradual correction of errors and leading to improved preservation of thin and connected structures. Overall, the experimental results demonstrate that CurvSegFlow achieves strong and consistent performance across a range of challenging conditions.  

In particular, our results highlight several key findings about CurvSegFlow. It consistently outperforms conventional encoder–decoder models for microtubule segmentation, especially under challenging conditions (Table~\ref{tab:complex}). The improvement mainly stems from better precision while maintaining competitive sensitivity, showing that the model reduces false positives without sacrificing true filament detection. The iterative vector field integration helps prevent error propagation across the image, common issue in single-pass decoders.  
Performance gains increase with structural complexity: on the MicSim\_FluoMT-Simple dataset improvements are modest (Table~\ref{tab:simple}), but on the MicSim\_FluoMT-Complex dataset, CurvSegFlow shows a clear advantage (Table~\ref{tab:complex}). This suggests that the flow-based approach excels in low-contrast, overlapping filaments, and noisy images, where pixel-wise classification often fails to maintain connectivity. Notably, the SynSeg pretrained model, which is dedicated to cytoskeleton segmentation, failed to achieve accurate segmentation (Table~\ref{tab:complex}, Figure \ref{fig:seg_higaki}d), underscoring the difficulties of segmenting microtubules in adverse scenarios and the need for advanced tools. 

CurvSegFlow also demonstrates robustness under limited supervision, as illustrated by its competitive performance on the MicReal\_FluoMT dataset, despite being trained on only 29 images (Table~\ref{tab:micreal}). Experiments with reduced data show a gradual performance decline rather than sudden failure, with AUC-ROC and specificity remaining high. Learning a transformation from noisy initialization to clean masks appears to regularize the model and stabilize predictions, even with few annotated images.   

Since we focus on curvilinear structures, we evaluated CurvSegFlow’s ability to preserve microtubule continuity and geometry by computing topology-aware metrics. The HD95 and clDice scores indicate that CurvSegFlow better preserves structural continuity and reduces fragmentation compared to other architectures, particularly in adverse scenarios (Tables~\ref{tab:complex}–\ref{tab:micreal}). These findings suggest that iterative flow-based refinement is well suited for capturing the geometry and topology of microtubules while progressively reducing uncertainty during reconstruction. Thus, it holds promise for segmenting other curvilinear structures beyond microtubules. 

On external curvilinear datasets, including retinal vessels, corneal nerve fibers, and coronary arteries (Tables~\ref{tab:drive}–\ref{tab:arcade}), CurvSegFlow remains competitive. While it does not always outperform specialized architectures in every metric, it consistently achieves strong values, indicating stable performance across different datasets. On the CORN-1 dataset, which features nerves in noisy backgound fluorescence, CurvSegFlow outperforms all other models across all metrics, confirming its effectiveness in adverse conditions (Table \ref{tab:corn1}). Its performance on the CHASE\_DB1 dataset, trained on only 20 images, further supports its efficacy in segmenting curvilinear structures despite limiting training data (Table \ref{tab:chasedb1}). Overall, this work demonstrates CurvSegFlow’s robustness across various curvilinear segmentation tasks without requiring architectural changes, its ability to perform well despite limiting training data, and its outperformance of other approaches in adverse conditions, such as high noise and low contrast.  

Despite its strong empirical performance, the proposed model has some limitations and leaves room for improvements. First, the iterative refinement introduces higher computational cost than single-pass models, although we show that as few as three Euler steps achieve competitive accuracy (Table \ref{tab:results_ts}). Second, while we quantitatively evaluate topology preservation and overlapping and demonstrate CurvSegFlow’s improvements in adverse scenarios, errors can still occur in extreme cases, which may be critical for some applications. It would thus be of interest to provide readouts of prediction quality to inform users of the prediction certainty level, particularly in regions where segmentation might be more challenging. Third, CurvSegFlow depends on annotations, requiring both high-quality annotations and a sufficient number of samples, although we show that 20 images can be sufficient. However, for some datasets, annotating even 20 images is time-consuming or unfeasible.

The present work and its limitations point to several promising future directions. First, incorporating uncertainty estimation could improve the handling of ambiguous regions. Interestingly, the iterative approach of Flow Matching naturally generates information about uncertainty at each reconstruction step, offering possibilities for developing uncertainty quantification methods and improving interpretability \citep{durasov2024enabling, parikh2025conditional, wu2026uncertainty}. For instance, the convergence rate at a given pixel could serve as a relevant indicator of uncertainty. Second, combining CurvSegFlow with semi-supervised or weakly supervised learning could further reduce the reliance on annotated data, a common bottleneck in deep learning. Third, the flow-based approach could be further improved by deriving theoretical bounds on topology preservation. This may benefit curvilinear segmentation in adverse conditions and extend applications to other fields, such as segmenting other thin structures (e.g. actin filaments) or detecting surface cracks. Last, adapting CurvSegFlow to 3D or time-lapse images would allow the mapping of networks extending in three dimensions or the modeling of filament dynamics. Such extentions could be highly relevant for biologists or medical professionals, as spatial or temporal inhomogeneities often reveal deregulations or diseases.  

Overall, this work highlights the potential of flow-based formulations – treating segmentation as structural reconstruction rather than pixel classification – as a general and flexible framework for biomedical image segmentation, particularly in scenarios where robustness, structural consistency, and uncertainty handling are critical. Future work will focus on better understanding and modeling uncertainty, extending the approach to three-dimensional and multimodal data, and improving computational efficiency for large-scale and real-time applications.

\section*{Acknowledgment}
This work was supported by the Agence Nationale de la Recherche (ANR-22-CE45-001601), by Campus France/ CNRST (PHC Toubkal 2024, n° 49945RE), the University of Rennes (Soutien aux collaborations internationales, 2024), the grant ERC-2022-SYG 101071583 from the European Research Council, and the grant 25-16671S from the Czech Science Foundation. The servers used for the computations were funded by the Brittany region (AAP PME 2018-2019 - Roboscope) and by the Agence Nationale de la Recherche (PRCE project SAMIC, ANR-19-CE45-0011). We thank Drs Guangshuo Ou and Zhengyang Guo for generously sharing the revised Higaki 2024 dataset, which was instrumental in evaluating our model’s performance. We thank Veronika Rumlova and Christina de Brito for technical assistance. We acknowledge institutional support from the CAS, Imaging Methods Core facility at BIOCEV and the CF Protein Production of the CIISB. We thank Pécréaux lab for its support. 

\printcredits


\bibliographystyle{cas-model2-names}

\bibliography{cas-refs}



\end{document}